\documentclass[12pt,draftcls,onecolumn]{IEEEtran}

\usepackage{graphicx,multicol}
\usepackage{psfrag,amsbsy,pstricks,amsmath,amsfonts,amssymb,cite}

\usepackage{booktabs}

\newcommand{\y}{\boldsymbol{y}}
\newcommand{\x}{\boldsymbol{x}}
\newcommand{\h}{\boldsymbol{h}}
\newcommand{\w}{\boldsymbol{w}}
\newcommand{\z}{\boldsymbol{z}}
\newcommand{\bvec}{\boldsymbol{u}}
\newcommand{\cvec}{\boldsymbol{c}}
\newcommand{\X}{\boldsymbol{X}}
\newcommand{\Hmat}{\boldsymbol{H}}

\newcommand{\herm}{^{\textrm{H}}}
\newcommand{\tra}{^{\textrm{T}}}
\newcommand{\mess}[2]{m_{#1 \to #2}}
\newcommand{\Mess}[1]{\boldsymbol{M}_{\textrm{#1}}}
\newcommand{\ness}[2]{n_{#1 \to #2}}
\newcommand{\Ness}[1]{\boldsymbol{N}_{\textrm{#1}}}
\newcommand{\trace}[1]{\textrm{Tr}\left\{ #1 \right\}}
\newcommand{\B}{\boldsymbol{\hat B}}
\newcommand{\C}{\boldsymbol{\hat C}}
\newcommand{\U}{\boldsymbol{U}}
\newcommand{\D}{\boldsymbol{\hat D}}
\newcommand{\covEst}[1]{\hat{\boldsymbol{\Sigma}}_{#1}}
\newcommand{\covMat}[1]{{\boldsymbol{\Sigma}}_{#1}}
\DeclareMathOperator{\argmax}{argmax}

\begin{document}
\title{Receiver Architectures for MIMO-OFDM Based on a Combined VMP-SP Algorithm}

\author{Carles Navarro Manch\'on, 
~Gunvor~E.~Kirkelund, 
 Erwin~Riegler, 
 Lars~P.~B.~Christensen, 
 Bernard~H.~Fleury. 
\thanks{This work has been submitted to the IEEE for possible publication. Copyright may be transferred without notice, after which this version may no longer be accessible.}
}


\maketitle

\begin{abstract}
Iterative information processing, either based on heuristics or
analytical frameworks, has been shown to be a very powerful tool for
the design of efficient, yet feasible, wireless receiver
architectures. Within this context, algorithms performing
message-passing on a probabilistic graph, such as the sum-product
(SP) and variational message passing (VMP) algorithms, have become
increasingly popular.

In this contribution, we apply a combined VMP-SP message-passing
technique to the design of receivers for MIMO-ODFM systems.
The message-passing equations of the combined scheme can be obtained
from the equations of the stationary points of a constrained
region-based free energy approximation. When applied to a MIMO-OFDM
probabilistic model, we obtain a generic receiver architecture
performing iterative channel weight and noise precision estimation,
equalization and data decoding. We show that this generic scheme can
be particularized to a variety of different receiver structures,
ranging from high-performance iterative structures to low complexity
receivers. This allows for a flexible design of the signal
processing specially tailored for the requirements of each specific
application. The numerical assessment of our solutions, based on
Monte Carlo simulations, corroborates the high performance of the
proposed algorithms and their superiority to heuristic approaches.

\end{abstract}


\IEEEpeerreviewmaketitle

\section{Introduction}

During the last two decades, wireless communication systems have
undergone a rapid and steep evolution. While old analog systems
mainly focused on providing voice communications, today's digital
systems offer a plethora of different services such as multimedia
communications, web browsing, audio and video streaming, etc. Along
with the growing variety of services offered, the amount of users
accessing them has also experienced a drastic increase. The
combination of applications requiring large amounts of data traffic
and high density of users, together with the scarceness of wireless
spectrum resources, dictates high spectral efficiency to be an
essential target in the design of modern wireless systems.

From a physical layer point of view, the emergence of multiple-input
multiple-output (MIMO) techniques~\cite{Gesbert2003} together with
the development of near-capacity-achieving channel codes, such as
turbo~\cite{Berrou1993} or low-density parity check
(LDPC)~\cite{Gallager1962} codes, have been the most remarkable
steps towards this goal. The use of multiple antennas allows for
increasing the theoretical capacity of a wireless channel linearly
with the minimum of the number of antenna elements at the
transmitter and at the receiver ends~\cite{Telatar1999}. Depending
on the specific MIMO technique employed, multiple antennas can be
used to exploit the number of degrees of freedom of a wireless
channel, its diversity or a mixture of both~\cite{Zheng2003}. The
combination with advanced channel codes enables transmission schemes
with unprecedented high spectral efficiency. However, in order to
realize in practice the performance predicted by theory, advanced
receiver architectures combining high performance channel
estimators, MIMO detectors and channel decoders are required.

Joint maximum likelihood (ML) receivers are prohibitively complex
for most modern communication systems, especially systems with high
MIMO order and concatenated codes. A wide-spread approach for the
design of suboptimal, yet efficient receiver architectures is to
separate the receiver into several individual blocks, each
performing a specific task: channel weight estimation, noise
estimation, interference cancellation, equalization or data decoding
are some examples. Inspired by the iterative decoding scheme of
turbo codes, some structures in which the different constituent
blocks exchange information in an iterative manner have been
proposed~\cite{Wang1999,Koetter2004,Loncar2004,Wehinger2006,Rossi2008a}.
In these receivers, each block is designed individually, and the way
it exchanges information with the other blocks is based on
heuristics. Consequently, while each block is designed to optimally
perform its task, the full receiver structure does not necessarily
optimize any global performance criterion. Nevertheless, these
structures have shown remarkably good performance at an affordable
complexity, while keeping a large degree of flexibility in their
design.


Motivated by the success of heuristic iterative approaches, a set of
formal frameworks for the design of algorithms performing iterative
information processing have arisen in recent years. Among these,
methods for variational Bayesian inference in probabilistic
models~\cite{Wainwright2008} have attracted much attention from the
communication research community in recent times. These frameworks
allow for the design of iterative algorithms based on the
optimization of a global cost function. Typically, they are derived
from the stationary points of a discrepancy measure between the
probability distribution that needs to be estimated and a postulated
auxiliary distribution, the latter distribution providing an
estimate of the former. The different frameworks differ on the
particular discrepancy measure selected and the restrictions applied
to the postulated auxiliary function. We especially highlight two
main approaches suggested so far in literature: belief propagation
(BP) and mean-field (MF) methods\footnote{Some authors, e.g. Winn
and Bishop~\cite{Winn2004,Bishop2006}, consider BP outside the
variational Bayesian framework, and usually use the term
\emph{variational} only in the context of MF-like approximations. We
use, however, the more general view proposed e.g.
in~\cite{Yedidia2005,Wainwright2008,Minka2005}, which considers BP
as another algorithm for variational Bayesian inference.}.

BP \cite{pearl1988} is a Bayesian inference framework applied to
graphical probabilistic models. In its message-passing form
--referred to as the sum-product (SP)
algorithm~\cite{Kschischang2001}-- messages are sent from one node
of the graphical model to neighboring nodes. The message computation
rules for the SP algorithm are obtained from the stationary points
of the Bethe free energy~\cite{Yedidia2005}. When the graphical
model representing the system is free of cycles, the SP algorithm
provides exact marginal distributions of the variables in the model.
When the graph has cycles, however, the algorithm outputs only an
approximation of the marginal distributions and it is, moreover, not
guaranteed to converge~\cite{Murphy99}. In most cases, nonetheless,
the obtained marginals are still a high quality approximation of the
exact distributions. BP and the SP algorithm have found widespread
application in the decoding of channel
codes~\cite{Kschischang2001,McEliece1998}, and have also been
proposed for the design of iterative receiver structures in wireless
communication
systems~\cite{Boutros2002,Loeliger2007,Colavolpe2005,Novak2008,Worthen2001}.
However, modifications of the original algorithm are required for
parameter estimation problems, such as channel estimation. This has
been solved by, e.g., combining the SP algorithm with the
expectation-maximization (EM) algorithm~\cite{Moon2004,Loeliger2007}
or approximating SP messages which are computationally untractable
with Gaussian messages~\cite{Novak2008b,Novak2009}.

MF approaches --proposed by Attias in~\cite{Attias1999} and
formulated as the variational Bayesian expectation-maximization
(VBEM) principle by Beal~\cite{Beal2003}-- are based on the
minimization of the Kullback-Leibler (KL)
divergence~\cite{Cover2006} between a postulated auxiliary function
and the distribution to be estimated. The minimization becomes
especially computationally tractable under the MF
approximation~\cite{Parisi1988}, in which the auxiliary function is
assumed to completely factorize with respect to the different
parameters. The obtained iterative algorithm guarantees convergence
in terms of the KL divergence, but convergence to the globally
optimum solution can only be guaranteed when the considered problem
has a unique single optimum. However, it has proven very useful in
the design of iterative receiver structures including channel
estimation, e.g., channel estimation and detection for GSM
systems~\cite{Christensen2006}, iterative multiuser channel
estimation, detection and decoding~\cite{Hu2008} or channel
estimation, interference cancellation and detection in OFDM
systems~\cite{Navarro2009a,Navarro2009b}. For other applications of
MF methods, see~\cite{Lin2007,Nissila2008,Zhang2008}.
Message-passing interpretations of this technique on probabilistic
graphs have also been proposed
in~\cite{Winn2004,Dauwels2007,Kirkelund2010} and are commonly
referred to as variational message-passing (VMP) techniques.

In this contribution, we apply a hybrid message-passing framework to
the design of iterative receivers 
in a MIMO-OFDM setup. This hybrid framework, recently proposed
in~\cite{Riegler2010,Riegler2011}, combines the SP and VMP
algorithms in a unified message-passing technique. Message updates
are obtained from the stationary points of a particular region-based
free energy approximation~\cite{Yedidia2005} of the probabilistic
system. Specifically, the combined framework allows for performing
VMP in parts of the graph and SP in others, thus
enabling a flexible, yet global, design.

From a MIMO-OFDM signal model, we derive a generic message-passing
receiver performing channel estimation, MIMO detection and channel
decoding in an iterative fashion. Channel estimation is not limited
to the estimation of channel weights, but also includes estimation
of the noise variance, which proves to be crucial for the operation
of the receiver. The application of a unified framework to the whole
receiver design unequivocally dictates the type of information that
should be exchanged by the individual constituents of the receiver
in the form of messages. This is in contrast to heuristic approaches
which, for instance, arbitrarily select a-posteriori or extrinsic
probabilities to be exchanged between the channel decoder and other
modules based on intuitive argumentation or trends observed by
simulation results~\cite{Wehinger2006,Rossi2008a}.

The generic messages derived can easily be particularized by
applying different assumptions and restrictions to the signal model
considered. Thus, our framework enables a highly scalable and
flexible design of the signal processing in the receiver. For
instance, applying the messages to only part of the factor graph
yields simplified architectures performing just a subset of the
receiver tasks; also, small modifications to the factor-graph lead
to different receiver structures with different performance and
computational complexity tradeoffs. These properties are illustrated
in our numerical evaluation, where the performance of a few selected
instances of our proposed receiver is assessed via Monte Carlo
simulations. The presented results demonstrate the high accuracy of
our approach, and its superiority to iterative receivers based on
heuristics.


The remainder of the paper is organized as follows. The signal model
of the MIMO-OFDM system considered is presented in
Section~\ref{Sec:SigMod}, followed by a brief review of the combined
message-passing framework proposed in~\cite{Riegler2010,Riegler2011}
in Section~\ref{Sec:MesPas}. In Section~\ref{Sec:GenRec}, the
generic messages to be exchanged in the factor-graph are derived,
and the performance of five different receivers obtained from the
generic derivation is tested in Section~\ref{Sec:Results}. Finally,
we draw some final conclusions in Section~\ref{Sec:Conc}.

\subsection{Notation}

Throughout the paper, lower-case boldface letters represent column
vectors, while upper-case boldface letters denote matrices;
$(\cdot)\tra$ and $(\cdot)\herm$ denote the transpose and
conjugate-transpose of a vector or matrix respectively; $\|\cdot\|$
denotes the Euclidian norm; $\boldsymbol{A}\otimes\boldsymbol{B}$
represents the Kronecker product of matrices $\boldsymbol{A}$ and
$\boldsymbol{B}$; $\boldsymbol{I}_N$ denotes the identity matrix of
dimension $N$. Moreover, $\log$ denotes the natural logarithm;
$f(x)\propto g(x)$ means that $f(x)$ is equal to $g(x)$ up to a
proportionality constant; $\langle f(x) \rangle_g$ denotes the
expectation of $f(x)$ over $g(x)$, i.e. $\langle f(x)
\rangle_g=\int_x f(x)g(x)dx$; $\mathcal{S}\backslash s$ denotes all
elements in the set $\mathcal{S}$ but $s$.

\section{Signal Model}
\begin{figure*}
\centering
\begin{psfrags}
\psfrag{pp}[c][c][1.1]{$+$} \psfrag{Pi}[c][c][1.1]{$\Pi$}
\psfrag{y}[c][c][1]{$\boldsymbol{y}$}
\psfrag{Chan1}[c][c][1]{$\boldsymbol{H}_1$}
\psfrag{Chanj}[c][c][1]{$\boldsymbol{H}_M$}
\psfrag{w}[c][c][1]{$\boldsymbol{w}$} \psfrag{yi}[c][c][1]{}
\psfrag{yj}[c][c][1]{} \psfrag{y1}[c][c][1]{}
\psfrag{b}[c][c][1]{$\boldsymbol{u}_1$}
\psfrag{bj}[c][c][1]{$\boldsymbol{u}_M$}
\psfrag{d}[c][c][1]{$\boldsymbol{x}_1$}
\psfrag{j}[c][c][1]{$\boldsymbol{x}_M$}
\psfrag{c}[c][c][1]{$\boldsymbol{c}_1$}
\psfrag{mi}[c][c][1]{$\boldsymbol{x}^{(p)}_1$}
\psfrag{m}[c][c][1]{$\boldsymbol{x}^{(d)}_1$}
\psfrag{mij}[c][c][1]{$\boldsymbol{x}^{(p)}_M$}
\psfrag{mj}[c][c][1]{$\boldsymbol{x}^{(d)}_M$}
\psfrag{cj}[c][c][1]{$\boldsymbol{c}_M$}
\psfrag{s}[c][c][1]{$\boldsymbol{c}^\pi_1$}
\psfrag{sj}[c][c][1]{$\boldsymbol{c}^\pi_M$}
\psfrag{Mul}[c][c][0.8]{Multiplexing}
\psfrag{Recf}[c][c][0.8]{\begin{minipage}{2.2cm}\footnotesize{Receiver
filter\\FFT+CP removal}\end{minipage}}
\psfrag{mod}[c][c][0.8]{Modulation} \psfrag{Enc}[c][c][0.8]{Encoder}
\psfrag{Pilot}[c][c][0.8]{\begin{minipage}{2cm}Pilot\\generation\end{minipage}}
\psfrag{Bit}[c][c][0.8]{}
\psfrag{m2}[c][c][0.8]{$\boldsymbol{m}_{\boldsymbol{h}_m\rightarrow
\boldsymbol{y}^p}$}
\psfrag{m3}[c][c][0.8]{$\boldsymbol{m}_{y\rightarrow x}$}
\psfrag{m4}[c][c][0.8]{$\boldsymbol{m}_{x\rightarrow y}$}
\psfrag{m5}[c][c][0.8]{$\boldsymbol{m}_{\pmb{\Lambda}^p\rightarrow
\boldsymbol{y}^p}$}
\psfrag{m6}[c][c][0.8]{$\boldsymbol{m}_{\boldsymbol{y}^p\rightarrow
\pmb{\Lambda}^p}$}
\includegraphics[width=\textwidth]{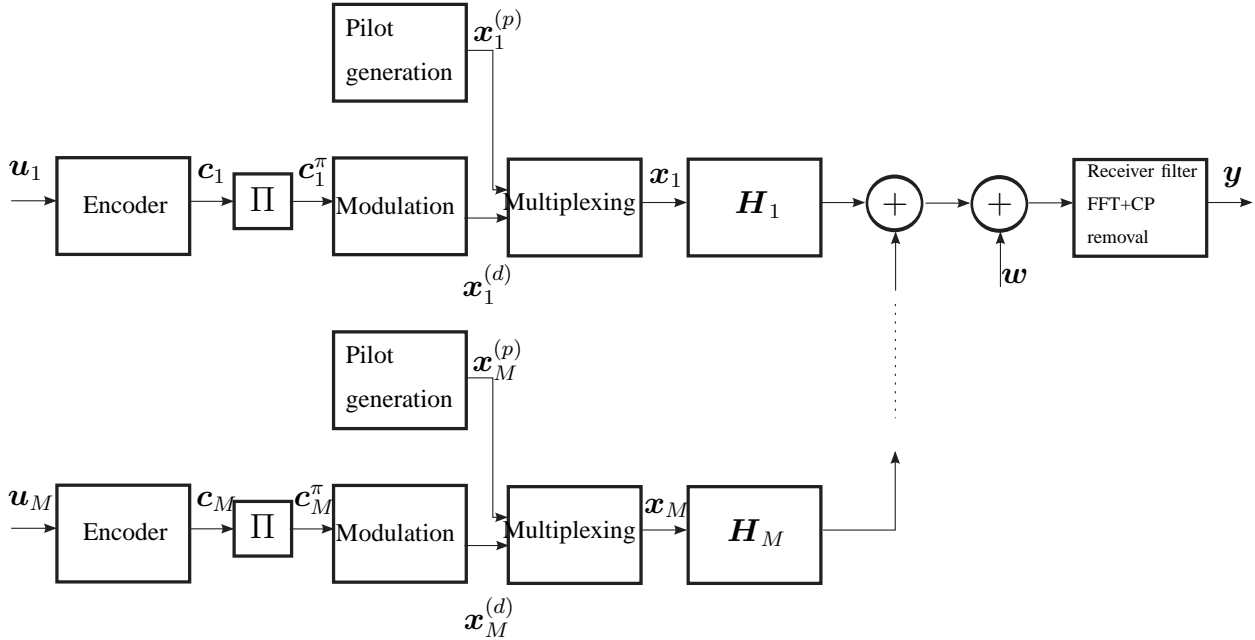}\\
\end{psfrags}
\caption{Block-diagram representation of the transmission model.}
\label{fig:Signalmodel}
\end{figure*}

In this section a multi-user signal model for MIMO-OFDM is derived.
The system is composed by $M$ synchronous transmitter chains and $N$
receiver antennas, as depicted in Fig.~\ref{fig:Signalmodel}. These
transmitters can represent different transmission branches of the
same physical transmitter, or physically separated transmitters at
different locations. For the $m$th transmitter, a finite sequence of
information bits $\boldsymbol{u}_m$ is encoded, yielding a sequence
of coded bits $\boldsymbol{c}_m$. After interleaving, the
interleaved coded bits $\boldsymbol{c}_m^{\pi}$ are complex
modulated, resulting in the vector $\boldsymbol{x}_m^{(d)}$ of
complex-modulated data symbols. Finally, the data symbols are
multiplexed with the pilot symbols $\boldsymbol{x}_m^{(p)}$, giving
the transmitted symbols
$\boldsymbol{x}_m=[x_m(1,1),\dots,x_m(K,1),\dots,x_m(1,L),\dots,
x_m(K,L)]^{\mathrm{T}}$, where $x_m(k,l)$ denotes the symbol sent by
the $m$th transmitter on the $k$th subcarrier of the $l$th OFDM
symbol of a frame. The transmitted symbols $\boldsymbol{x}_m$ are
then OFDM modulated using an IFFT and the insertion of a cyclic
prefix.

The signal is transmitted through a wide-sense stationary
uncorrelated scattering (WSSUS) channel. The channel impulse
response from transmitter $m$ to receiver $n$ during the
transmission of the $l$th OFDM symbol $l$ can be described by
\begin{equation}\label{eq:CIR}
g_{nm}(l,\tau)=\sum_{i=1}^{I_{nm}}
\alpha_{nm}^{(i)}(l)\delta(\tau-\tau_{nm}^{(i)})
\end{equation}
where $\alpha_{nm}^{(i)}$ and $\tau_{nm}^{(i)}$ are respectively the
complex gain and delay of the $i^\mathrm{th}$ multipath component
and $I_{nm}$ is the number of multipath components. We assume that
the channel response is static over the duration of an OFDM symbol,
but changes from one OFDM symbol to the next. Also, the maximum
delay of each wireless link $\tau_{nm}^{(I_{nm})} $ is assumed to be
smaller than the duration of the OFDM cyclic prefix\footnote{We
assume without loss of generality that the delays $\tau_{nm}^{(i)}$
are ordered in increasing order, i.e.
$\tau_{nm}^{(i+1)}\geq\tau_{nm}^{(i)}$.}, so that no inter-symbol
interference (ISI) degrades the transmission. From (\ref{eq:CIR}),
the sample of the channel frequency response at the $k$th subcarrier
of the $l$th OFDM symbol is found to be:
\begin{equation*}
h_{nm}(k,l)=\sum_{i=1}^{I_{nm}}\alpha_{nm}^{(i)}(l)e^{-j2\pi
k\Delta_f\tau_{nm}^{(i)}}.
\end{equation*}
In this expression, $\Delta_f$ denotes the OFDM subcarrier spacing.

At the receiver, the signal is OFDM demodulated by discarding the
cyclic prefix and applying an FFT on the received samples. Under the
previously stated assumptions that the channel is block fading and
the maximum delays are smaller than the duration of the cyclic
prefix, the signal received at the $n$th receive antenna on the
$k$th subcarrier of the $l$th OFDM symbol reads
\begin{equation}\label{eq:rec_signal_scalar}
y_n(k,l)=\sum_{m=1}^{M}h_{nm}(k,l)x_m(k,l)+w_n(k,l),\quad
\begin{array}{l} n=1,\dots,N,\\ k=1,\dots,K,\\
l=1,\dots,L,\end{array}
\end{equation}
with $w_n(k,l)$ denoting zero-mean additive complex white Gaussian
noise (AWGN) with variance $\lambda^{-1}$. The equations in
\eqref{eq:rec_signal_scalar} can be recast in a matrix-vector
notation as
\begin{equation}\label{eq:rec_signal_matrix}
\boldsymbol{y}=\sum_{m=1}^{M}\boldsymbol{X}_m\boldsymbol{h}_m+\boldsymbol{w}
=\sum_{m=1}^{M}\boldsymbol{H}_m\boldsymbol{x}_m+\boldsymbol{w}
\end{equation}
where $\boldsymbol{y}=[\boldsymbol{y}_1\tra,\dots,
\boldsymbol{y}_N\tra]\tra$, with $\boldsymbol{y}_n=[y_n(1,1),\dots,
y_n(K,1),\dots,y_n(1,L),\dots,y_n(K,L)]^T$ denoting the received
signal at the $n$th receive antenna for a frame of $K$ subcarriers
and $L$
OFDM symbols. Additionally, 
$\boldsymbol{h}_m=[\boldsymbol{h}_{1m}^{\mathrm{T}},\dots,\boldsymbol{h}_{Nm}^{\mathrm{T}}]^{\mathrm{T}}$,
$\boldsymbol{X}_m =
\boldsymbol{I}_N\otimes\mathrm{diag}\{\boldsymbol{x}_m\}$,
$\boldsymbol{H}_m = [\mathrm{diag}\{\boldsymbol{h}_{1m}\}, \dots,
\mathrm{diag}\{\boldsymbol{h}_{Nm}\}]^{\mathrm{T}}$ and
$\boldsymbol{h}_{nm} =
[h_{nm}(1,1),\dots,h_{nm}(K,1),\dots,h_{n,m}(1,L),\dots,
h_{nm}(K,L)]^\mathrm{T}$. Equation \eqref{eq:rec_signal_matrix} can
be further compressed as
\begin{equation*}
\y=\X\h+\w=\Hmat\x+\w
\end{equation*}
where $\boldsymbol{x}=[\boldsymbol{x}_1\tra, \dots,
\boldsymbol{x}_M\tra]\tra$, $\boldsymbol{h}=[\boldsymbol{h}_1\tra,
\dots ,\boldsymbol{h}_M\tra]\tra$, $\X=[\X_1,\dots,\X_M]$ and
$\Hmat=[\Hmat_1,\dots,\Hmat_M]$.
%
\label{Sec:SigMod}

\section{Message Passing Techniques}
\label{Sec:MesPas}
In this section, we briefly introduce message-passing techniques on
factor graphs. First, we define the concept of factor graph on a
probabilistic model, followed by the description of two standard
message-passing schemes: the sum-product (SP)
algorithm~\cite{Kschischang2001} and the variational message-passing
(VMP) algorithm~\cite{Winn2004}. Finally, we show how to combine
both algorithms to perform hybrid VMP and SP message passing in a
factor graph~\cite{Riegler2010}.

\subsection{Factor Graphs for Probabilistic Models}

Let $p(\z)$ be the probability density function (pdf) of a vector
$\z$ of random variables $z_i$ ($i\in \mathcal{I}$) which factorizes
according to
\begin{equation}
p(\z)=\frac{1}{Z}\prod_{a\in\mathcal{A}}f_a(\z_a)
\label{eq:prob_mod}
\end{equation}
where $\z_a=(z_i|i\in\mathcal{N}(a))\tra$ with
$\mathcal{N}(a)\subseteq\mathcal{I}$ for all $a\in\mathcal{A}$ and
$Z=\int_{\z}\prod_{a\in\mathcal{A}}f_a(\z_a)d\z$ is a normalization
constant. We also define
$\mathcal{N}(i)\triangleq\{a\in\mathcal{A}|i\in\mathcal{N}(a)\}$ for
all $i\in\mathcal{I}$. Similarly,
$\mathcal{N}(a)=\{i\in\mathcal{I}|a\in\mathcal{N}(i)\}$ for all
$a\in\mathcal{A}$. The above factorization can be graphically
represented by means of a factor graph~\cite{Kschischang2001}. A
factor graph\footnote{We will use Tanner factor
graphs~\cite{Kschischang2001} throughout this article} is a
bipartite graph having a variable node $i$ (typically represented by
a circle) for each variable $z_i$, $i\in \mathcal{I}$ and a factor
node $a$ (represented by a square) for each factor $f_a$,
$a\in\mathcal{A}$. An edge connects a variable node $i$ to a factor
node $a$ if, and only if, the variable $z_i$ is an argument of the
factor function $f_a$. The set $\mathcal{N}(i)$ contains all factor
nodes connected to a variable node $i\in\mathcal{I}$ and
$\mathcal{N}(a)$ is the set of all variable nodes connected to a
factor node $a\in\mathcal{A}$.

Factor graphs provide a compact and intuitive representation of the
statistical dependencies among the random variables in a
probabilistic model. Furthermore, they enable the design of a class
of iterative signal processing algorithms which are based on the
nodes of the graph iteratively exchanging information (messages)
with their neighbors (connected nodes). This class of algorithms has
been coined \emph{message-passing} techniques, and in the following
we will describe the two instances of these techniques which have
been most widely applied to signal processing for communication
systems: the SP and VMP algorithms.

\subsection{The Sum-Product Algorithm}

The SP algorithm is a message-passing algorithm that computes the
exact marginal distributions $p_i(z_i)$ of the variables $z_i$
associated to the joint distribution $p(\z)$ for tree-shaped factor
graphs. When the factor graph does not have a tree structure, the
outcome of the algorithm is only an approximation of the true
marginal, and the approximate marginals $b_i(z_i)\approx p_i(z_i)$
are called beliefs. The message-passing algorithm is derived from
the equations of the stationary points of the constrained Bethe free
energy ~\cite{Yedidia2005}.

The algorithm operates iteratively by exchanging messages from
variable nodes to factor nodes and vice-versa. The message
computation rules for the SP algorithm read
\begin{align*}
\mess{a}{i}(z_i) & = d_a\langle f_a(\z_a) \rangle_{\prod_{j\in
\mathcal{N}(a)\backslash i}\ness{j}{a}}, \quad \forall a\in\mathcal{A}, i\in\mathcal{N}(a)\\
\ness{i}{a}(z_i) & = \prod_{c\in\mathcal{N}(i)\backslash a}
\mess{c}{i}(z_i), \quad \forall i\in\mathcal{I},a\in\mathcal{N}(i)
\end{align*}
where $d_a$ ($a\in\mathcal{A}$) are positive constants ensuring that
the beliefs are normalized to one. Often the constants $d_a$ need
not be calculated explicitly, and it is enough to normalize the
beliefs after convergence of the algorithm (see~\cite{Riegler2011}
for more details on normalization issues). We use the notation
$\ness{(\cdot)}{(\cdot)}$ for output messages from a variable node
to a factor node and $\mess{(\cdot)}{(\cdot)}$ for input messages
from a factor node to a variable node. This convention will be kept
through the rest of the paper, also for other message-passing
schemes.

The variables' beliefs can be calculated at any point during the
iterative algorithm as
\begin{equation*}
b_i(z_i)=\prod_{a\in\mathcal{N}(i)}\mess{a}{i}(z_i) \quad\forall
i\in\mathcal{I}.
\end{equation*}

The SP algorithm acquired great popularity through its application
to iterative decoding of, among others, turbo codes and LDPC codes,
and has since then been used for the design of many iterative
algorithms in a wide variety of fields~\cite{Loeliger2007}.

\subsection{The Variational Message-Passing Algorithm}

The VMP algorithm is an alternative message-passing technique which
is derived based on the minimization of the variational free energy
subject to the mean-field approximation constraint on the beliefs.
While it does not guarantee the computation of exact marginals (even
for tree-shaped graphs), its convergence is guaranteed by ensuring
that the variational free energy of the computed beliefs is
non-increasing at each step of the algorithm~\cite{Yedidia2005}.

The operation of the VMP algorithm is analogous to the SP algorithm;
the message computation rules read
\begin{align}
\mess{a}{i}(z_i) & = \exp\langle \log f_a(\z_a) \rangle_{\prod_{j\in
\mathcal{N}(a)\backslash i}\ness{j}{a}}, \quad \forall a\in\mathcal{A}, i\in\mathcal{N}(a)\label{eq:VMP_m_mes}\\
\ness{i}{a}(z_i) & = e_i\prod_{c\in\mathcal{N}(i)}
\mess{c}{i}(z_i)\quad \forall
i\in\mathcal{I},a\in\mathcal{N}(i)\label{eq:VMP_n_mes}
\end{align}
where $e_i$ ($i\in\mathcal{I}$) are positive constants ensuring that
$\ness{i}{a}$ are normalized. As in the SP algorihtm, the beliefs
can be obtained as
\begin{equation*}
b_i(z_i)=e_i\prod_{c\in\mathcal{N}(i)}\mess{c}{i}(z_i)=\ness{i}{a}(z_i)\quad\forall
i\in\mathcal{I},a\in\mathcal{N}(i).
\end{equation*}

The VMP algorithm has recently attracted the attention of the
wireless communication research community due to its suitability for
conjugate-exponential probabilistic models~\cite{Winn2004}. The
computation rule for input messages from factor to variable nodes
allows for the obtention of closed-form expressions in many cases in
which the SP algorithm typically requires some type of numerical
approximation.

It is shown in~\cite{Riegler2011} that a message-passing
interpretation of the EM algorithm can be obtained from the VMP
algorithm. Assume that for a certain subset of variables $z_i$,
$i\in\mathcal{E}\subseteq\mathcal{I}$ we want to apply an EM update
while still using VMP for the rest of variables. To do so, the
beliefs $b_i$ are restricted to fulfill the constraint $b_i(z_i) =
\delta(z_i-\tilde z_i)$ for all $i\in\mathcal{E}$ additionally to
the mean-field factorization and normalization constraints.
Minimizing the variational free energy subject to these conditions
leads to a message passing algorithm identical to the one described
in \eqref{eq:VMP_m_mes} and \eqref{eq:VMP_n_mes} except that the
messages $\ness{i}{a}$ for all $i\in\mathcal{E}$ and
$a\in\mathcal{N}(i)$ are replaced by
\begin{equation}
\ness{i}{a}(z_i)=\delta(z_i-\tilde z_i)\qquad
\mathrm{with}\quad\tilde z_i =
\argmax_{\z_i}\left(\prod_{a\in\mathcal{N}(i)}\mess{a}{i}(z_i)\right).\label{eq:EM_mes}
\end{equation}

\subsection{Combined VMP-SP Algorithm}\label{Sec:SPA-VMP}

As stated previously in this section, the VMP and the SP algorithms
are two message-passing techniques suitable for different types of
models. While SP is especially suitable in models with deterministic
factor nodes, e.g. code or modulation constraints, VMP has the
advantage of yielding closed-form computationally tractable
expressions in conjugate-exponential models, as are found in channel
weight estimation and noise variance estimation problems. Based on
these facts, it seems natural to try to combine the two methods in a
unified scheme capable of preserving the advantages of both.

A combined message-passing scheme based on the SP and VMP algorithms
was recently proposed in~\cite{Riegler2010,Riegler2011}. This hybrid
technique is based on splitting the factor graph into two different
parts: a VMP part and a SP part. To do this, part of the factor
nodes are assigned to the VMP set ($\mathcal{A}_{\mathrm{VMP}}$) and
the rest are assigned to the SP set ($\mathcal{A}_{\mathrm{SP}}$).
Given this classification, we can express the probabilistic model in
\eqref{eq:prob_mod} as
\begin{equation*}
p(\z)=\frac{1}{Z}\overbrace{\prod_{a\in\mathcal{A}_{\mathrm{VMP}}}f_a(\z_a)}^{\mathrm{VMP
part}}\overbrace{\prod_{c\in\mathcal{A}_{\mathrm{SP}}}f_c(\z_c)}^{\mathrm{SP
part}}
\end{equation*}
where
$\mathcal{A}_{\mathrm{VMP}}\cup\mathcal{A}_{\mathrm{SP}}=\mathcal{A}$
and
$\mathcal{A}_{\mathrm{VMP}}\cap\mathcal{A}_{\mathrm{SP}}=\emptyset$.
By applying the Bethe approximation to the SP part and the
mean-field approximation on the VMP part, 
a new message-passing scheme is derived from the stationary points
of the region-based free energy~\cite{Riegler2010,Riegler2011}. The
message computation rules for this algorithm read
\begin{align}
\mess{a}{i}^{\mathrm{VMP}}(z_i) & = \exp\langle \log f_a(\z_a)
\rangle_{\prod_{j\in
\mathcal{N}(a)\backslash i}\ness{j}{a}}, \quad \forall a\in\mathcal{A}_{\mathrm{VMP}}, i\in\mathcal{N}(a)\label{eq:VMP_mes}\\
\mess{a}{i}^{\mathrm{SP}}(z_i) & = d_a\langle f_a(\z_a)
\rangle_{\prod_{j\in
\mathcal{N}(a)\backslash i}\ness{j}{a}}, \quad \forall a\in\mathcal{A}_{\mathrm{SP}}, i\in\mathcal{N}(a)\label{eq:SPA_mes}\\
\ness{i}{a}(z_i) & = e_i
\prod_{c\in\mathcal{N}(i)\cap\mathcal{A}_{\mathrm{VMP}}}
\mess{c}{i}^{\mathrm{VMP}}(z_i)\prod_{c\in\mathcal{N}(i)\cap\mathcal{A}_{\mathrm{SP}}\backslash
a} \mess{c}{i}^{\mathrm{SP}}(z_i)\quad \forall
i\in\mathcal{I},a\in\mathcal{N}(i)\label{eq:n_mes}
\end{align}
where, again, $d_a$ and $e_i$ are positive constants ensuring
normalized beliefs. The computation rules for messages outgoing
factor nodes are preserved: for factor nodes in the VMP part
($a\in\mathcal{A}_{\mathrm{VMP}}$) the messages are computed using
\eqref{eq:VMP_mes} as in standard VMP; for factor nodes in the SP
part ($a\in\mathcal{A}_{\mathrm{SP}}$) the messages are computed via
\eqref{eq:SPA_mes}, which corresponds to a standard SP message. A
message from a variable node $i$ to a factor node $a$ is computed as
a VMP message when $a\in\mathcal{A}_{\mathrm{VMP}}$ and as a SP
message when $a\in\mathcal{A}_{\mathrm{SP}}$, as can be deduced from
$\eqref{eq:n_mes}$.

As with the VMP and SP algorithms, the beliefs of the variables can
be retrieved at any stage of the algorithm as
\begin{equation*}
b_i(z_i)=e_i\prod_{a\in\mathcal{N}(i)\cap\mathcal{A}_{\mathrm{VMP}}}\mess{a}{i}^{\mathrm{VMP}}(z_i)
\prod_{a\in\mathcal{N}(i)\cap\mathcal{A}_{\mathrm{SP}}}\mess{a}{i}^{\mathrm{SP}}(z_i)\quad
\forall i\in\mathcal{I}.
\end{equation*}

Note that we can apply the EM restriction to the belief of variables
$z_i$ which are only connected to VMP factors (i.e.
$\mathcal{N}(i)\cap\mathcal{A}_{\mathrm{SP}}=\emptyset$). In that
case, the message update rules remain the same except that the
message $\ness{i}{a}$ in \eqref{eq:n_mes} is replaced by
\eqref{eq:EM_mes} for the selected variables.

\section{MIMO-OFDM Receiver Based on Combined VMP-SPA} \label{Sec:GenRec}
In this section, we present a generic iterative receiver for
MIMO-OFDM systems based on the mixed VMP and SP message-passing
strategy outlined in Section~\ref{Sec:SPA-VMP}.
Recalling the signal model presented in Section~\ref{Sec:SigMod}, we
can now postulate the probabilistic model to which we will apply the
combined VMP-SP technique. In our case, we identify the observation
to be the received signal vector $\y$. As unknown parameters, we
include the vector of information bits
$\bvec=[\bvec_1\tra,\dots,\bvec_M\tra]\tra$, the vector of coded
bits $\cvec=[\cvec_1\tra,\dots,\cvec_M\tra]\tra$, the vector of
modulated symbols $\x=[\x_1,\dots,\x_M]\tra$, the vector of complex
channel weights $\h=[\h_1,\dots,\h_M]\tra$
and the AWGN precision $\lambda$. 
The system function of our model is the joint pdf of all parameters,
which can be factorized as
\begin{equation}
p(\bvec,\cvec,\x,\h,\lambda,\y)=\underbrace{p(\y|\h,\x,\lambda)}_{f_{\mathrm{O}}}\underbrace{p(\h)}_{f_{\mathrm{C}}}\underbrace{p(\lambda)}_{f_{\mathrm{N}}}\underbrace{p(\x,\cvec,\bvec)}_{f_{\mathrm{M}}}\label{eq:syst_func}
\end{equation}
where we have chosen to group the factors on the right-hand side
into four functions. Factor
$f_{\mathrm{O}}(\y,\h,\x,\lambda)\triangleq p(\y|\h,\x,\lambda)$
denotes the likelihood of the channel weights $\h$, the noise
precision $\lambda$ and the transmitted symbols $\x$ given the
observation $\y$. Factor $f_{\mathrm{C}}(\h)\triangleq p(\h)$
contains the assumed prior model of the channel weights, which is
relevant for channel weight estimation. Function
$f_{\mathrm{N}}(\lambda)\triangleq p(\lambda)$, likewise, contains
the assumed prior model for the noise precision parameter $\lambda$
which defines how estimation of the noise precision is done.
Finally, function $f_{\mathrm{M}}(\x,\cvec,\bvec)\triangleq
p(\x,\cvec,\bvec)$ denotes the modulation and code constraints. Note
that further factorization of the factors in \eqref{eq:syst_func} is
possible and will, in fact, be used later in this section.

\begin{figure}
\centering
\begin{psfrags}
\psfrag{fac}[c][c][1]{$f_{\mathrm{O}}$}
\psfrag{m1}[c][c][1]{$\boldsymbol{M}_{\mathrm{M}}$}
\psfrag{m2}[c][c][1]{$\boldsymbol{M}_{\mathrm{N}}$}
\psfrag{m3}[c][c][1]{$\boldsymbol{N}_{\mathrm{N}}$}
\psfrag{m4}[c][c][1]{$\boldsymbol{M}_{\mathrm{C}}$}
\psfrag{m5}[c][c][1]{$\boldsymbol{N}_{\mathrm{C}}$}
\psfrag{m6}[c][c][1]{$\boldsymbol{N}_{\mathrm{M}}$}
\psfrag{cloud1}[c][c][1]{Modulation and Coding}
\psfrag{cloud2}[c][c][1]{Noise Precision}
\psfrag{cloud3}[c][c][1]{Channel Weights}
\includegraphics[width=0.8\textwidth]{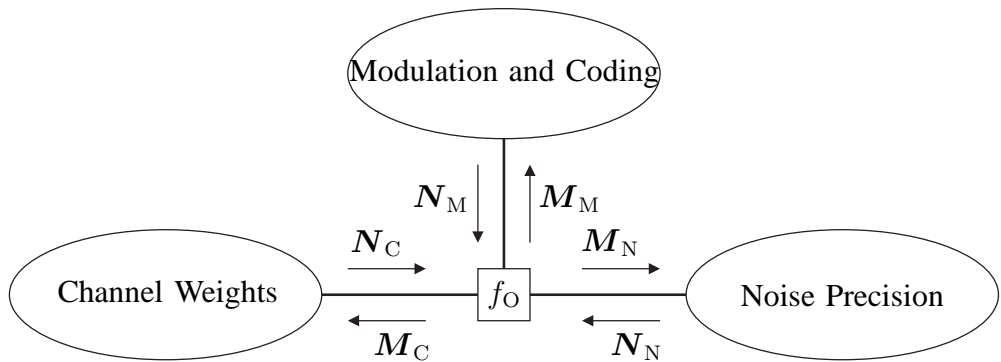}
\end{psfrags}
\caption{Generic factor graph of the receiver.}
\label{fig:FacGraphGen}
\end{figure}

A schematic factor-graph-like representation of the model in
\eqref{eq:syst_func} is depicted in Fig.~\ref{fig:FacGraphGen}. The
observation factor node $f_{\mathrm{O}}$ is connected to three
ovals: channel weights, noise precision and modulation and coding.
Each of the ovals represents a subgraph corresponding to factors
$f_{\mathrm{C}}$, $f_{\mathrm{N}}$ and $f_{\mathrm{M}}$ in
\eqref{eq:syst_func}. The three subgraphs are connected to
$f_{\mathrm{O}}$, which reads
\begin{align*}
f_{\mathrm{O}}(\y,\x,\h,\lambda) & \propto
\lambda^{KNL}\exp\left\{-\lambda\|\y -
\X\h\|^2\right\}=\lambda^{KNL}\exp\left\{-\lambda\|\y -
\Hmat\x\|^2\right\}.
\end{align*}
Each of the subgraphs in Fig.~\ref{fig:FacGraphGen} will be detailed
in the remainder of this section. For now, we define the sets
$\mathcal{A}_{\mathrm{C}}$, $\mathcal{A}_{\mathrm{N}}$ and
$\mathcal{A}_{\mathrm{M}}$ as the set of factor nodes inside the
channel weights, noise precision and modulation and coding subgraphs
respectively. Likewise, we define the sets
$\mathcal{I}_{\mathrm{C}}$, $\mathcal{I}_{\mathrm{N}}$ and
$\mathcal{I}_{\mathrm{M}}$ as the set of variable nodes inside the
channel weights, noise precision and modulation and coding subgraphs
respectively. With these definitions, the set of all factor nodes in
the graph is given by\footnote{With a slight abuse of notation, from
this point on we use the names of functions and variables as indices
of the sets $\mathcal{A}$ and $\mathcal{I}$ respectively.}
\begin{equation*}
\mathcal{A}=\{f_{\mathrm{O}}\}\cup\mathcal{A}_{\mathrm{C}}\cup\mathcal{A}_{\mathrm{N}}\cup\mathcal{A}_{\mathrm{M}},
\end{equation*}
and the set of all variable nodes reads
\begin{equation*}
\mathcal{I}=\mathcal{I}_{\mathrm{C}}\cup\mathcal{I}_{\mathrm{N}}\cup\mathcal{I}_{\mathrm{M}}.
\end{equation*}
From the observation factor node $f_{\mathrm{O}}$, sets of messages
$\Mess{C}$, $\Mess{N}$ and $\Mess{M}$ are sent to the respective
subgraphs. These sets are composed of individual messages
$\mess{f_{\mathrm{O}}}{z}$, $z\in\mathcal{I}$. The specific
composition of the sets of messages depends on the exact
configuration of variable and factor nodes of the corresponding
subgraph, which will be described later in the section. After
processing is completed at each subgraph, sets of messages
$\Ness{C}$, $\Ness{N}$ and $\Ness{M}$, which correspond to the
updated estimates of the channel weights, the noise precision and
the transmitted symbols respectively, are send back to
$f_{\mathrm{O}}$.

In order to apply the combined VMP-SP algorithm, we need to define
which factor nodes are assigned to the VMP set
$\mathcal{A}_{\mathrm{VMP}}$ and which are assigned to the SP set
$\mathcal{A}_{\mathrm{SP}}$. We select the following splitting:
\begin{align*}
\mathcal{A}_{\mathrm{VMP}} \triangleq &
\{f_{\mathrm{O}}\}\cup\mathcal{A}_{\mathrm{C}}\cup\mathcal{A}_{\mathrm{N}}
\\ \mathcal{A}_{\mathrm{SP}} \triangleq &
\mathcal{A}_{\mathrm{M}}
\end{align*}
i.e. the observation factor node and all factors in the channel
weight and noise precision subgraphs are assigned to the VMP set,
and all factor nodes in the modulation and coding subgraph are
assigned to the SP set.

In the remainder of this section, we will present the details of
each of the subgraphs, with several alternative factor-graph
representations yielding different message-passing configurations.
The performance of the individual receiver structures obtained will
be evaluated and compared in Section~\ref{Sec:Results}.

\subsection{Noise Precision Subgraph}\label{Sec:NPE}

\begin{figure}
\centering
\begin{psfrags}
\psfrag{fac}[c][c][1]{$f_{\mathrm{O}}$}
\psfrag{var}[c][c][1]{$\lambda$}
\psfrag{fac2}[c][c][1]{$f_{\mathrm{N}}$}
\psfrag{m1}[c][c][1]{$\boldsymbol{M}_{\mathrm{M}}$}
\psfrag{m2}[c][c][1]{$\boldsymbol{M}_{\mathrm{N}}$}
\psfrag{m3}[c][c][1]{$\boldsymbol{N}_{\mathrm{N}}$}
\psfrag{m4}[c][c][1]{$\boldsymbol{M}_{\mathrm{C}}$}
\psfrag{m5}[c][c][1]{$\boldsymbol{N}_{\mathrm{C}}$}
\psfrag{m6}[c][c][1]{$\boldsymbol{N}_{\mathrm{M}}$}
\psfrag{m7}[c][c][1]{$m_{f_{\mathrm{O}}\to \lambda}$}
\psfrag{m8}[c][c][1]{$n_{\lambda\to f_{\mathrm{O}} }$}
\psfrag{m9}[c][c][1]{$m_{f_{\mathrm{N}}\to \lambda }$}
\psfrag{cloud2}[c][c][1]{Noise Precision}
\includegraphics[width=0.8\textwidth]{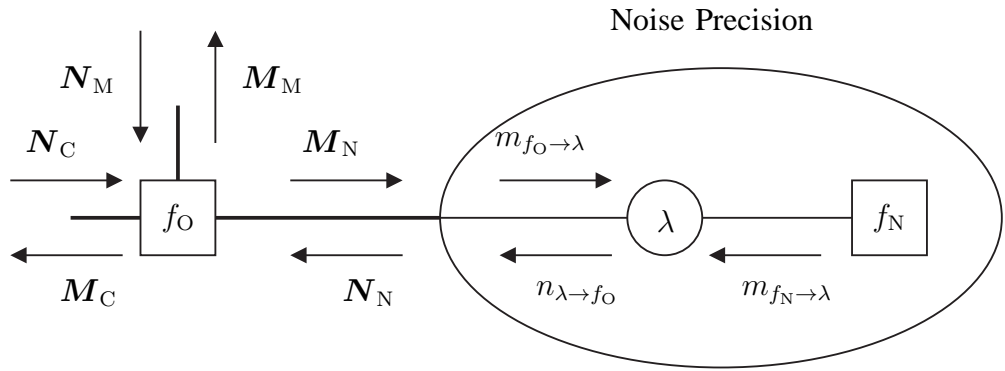}
\end{psfrags}
\caption{Subgraph corresponding to the noise precision prior
model.}\label{fig:FacGraphNPE}
\end{figure}

The noise precision subgraph is the graphical representation of
$f_{\mathrm{N}}$ in \eqref{eq:syst_func}, which we specify now as
\begin{equation*}
f_{\mathrm{N}}(\lambda)\triangleq p(\lambda)
\end{equation*}
where $p(\lambda)$ denotes the prior distribution of $\lambda$. With
this, we can now specify the sets
\begin{align*}
\mathcal{A}_{\mathrm{N}}= & \{f_{\mathrm{N}}\}\\
\mathcal{I}_{\mathrm{N}}= & \{\lambda\}.
\end{align*}
The factor graph representation of the subgraph is depicted in
Fig.~\ref{fig:FacGraphNPE}. It only consists of the variable node
$\lambda$ and the factor node $f_{\mathrm{N}}$. Since there is only
one variable node connected to $f_{\mathrm{O}}$, the set of messages
$\Mess{N}$ reduces to $\Mess{N}=\{\mess{f_{\mathrm{O}}}{\lambda}\}$.
Analogously, $\Ness{N}=\{\ness{\lambda}{f_{\mathrm{O}}}\}$.

According to the message-computation rules given in
Section~\ref{Sec:MesPas}, the message transmitted from
$f_{\mathrm{O}}$ to $\lambda$ is calculated as
\begin{equation}
\mess{f_{\mathrm{O}}}{\lambda}(\lambda) = \exp\left\{\langle\log
f_{\mathrm{O}}(\y,\x,\h,\lambda)\rangle_{\Ness{C}\Ness{M}}\right\} =
\lambda^{KLN}\exp\left\{ -\lambda A \right\}\label{eq:mess_y_lambda}
\end{equation}
with
\begin{equation*}
A = \|\y - \hat \X\hat\h\|^2 +\trace{\B\herm\C\B +
\B\herm\hat\Hmat\herm\hat\Hmat\B}+\trace{\hat\X\covEst{\h}\hat\X\herm}.
\end{equation*}
In the above expression, $\hat\h=\langle\h\rangle_{\Ness{C}}$,
$\hat\Hmat=\langle\Hmat\rangle_{\Ness{C}}$,
$\hat\x=\langle\x\rangle_{\Ness{M}}$,
$\hat\X=\langle\X\rangle_{\Ness{M}}$ are the means of $\h$, $\Hmat$,
$\x$ and $\X$ respectively taken with respect to the channel weights
and modulation and coding output messages. Moreover,
$\covEst{\h}=\langle\h\h\herm\rangle_{\Ness{C}}-\hat\h\hat\h\herm$,
and
$\C=\langle\Hmat\herm\Hmat\rangle_{\Ness{M}}-\hat\Hmat\herm\hat\Hmat$.
Finally, $\B=\U\boldsymbol{\Lambda}^{1/2}$ where
$\boldsymbol{\Lambda}$ is the diagonal matrix of eigenvalues and
$\U$ is the matrix containing the eigenvectors of
$\covEst{\x}=\langle\x\x\herm\rangle_{\Ness{M}}-\hat\x\hat\x\herm$,
i.e. $\covEst{\x}=\U\boldsymbol{\Lambda}\U\herm$.

The message in (\ref{eq:mess_y_lambda}) is proportional to the pdf
of a complex central Wishart distribution of dimension 1, $KLN+1$
degrees of freedom and associated covariance
$A^{-1}$~\cite{Tauge1994}. We select the prior pdf $p(\lambda)$ to
be conjugate, i.e., a complex Wishart. This yields the message
\begin{equation*}
\mess{f_{\mathrm{N}}}{\lambda}(\lambda) = p(\lambda) \propto
\lambda^{a-1}\exp\{-\lambda A_{\textrm{prior}}\}.
\end{equation*}
Given the two incoming messages $\mess{f_{\mathrm{N}}}{\lambda}$ and
$\mess{f_{\mathrm{O}}}{\lambda}$, the outgoing message from
$\lambda$ is also proportional to a complex Wishart pdf
\begin{equation*}
\ness{\lambda}{f_{\mathrm{O}}}(\lambda) \propto
\mess{f_{\mathrm{N}}}{\lambda}(\lambda)\mess{f_{\mathrm{O}}}{\lambda}(\lambda)
\propto \lambda^{KLN+a-1}\exp\{-\lambda(A+A_{\textrm{prior}})\}.
\end{equation*}
Since usually no prior information on the noise precision is
available at the receiver, we select $p(\lambda)$ non-informative
with parameters $a=0$ and $A_{\textrm{prior}}=0$. With this choice,
the mean of $\lambda$ with respect to $\Ness{N}$ reads
\begin{equation}
\hat\lambda=\langle \lambda \rangle_{\Ness{N}} = \frac{KLN}{A}.
\end{equation}
Note that the above update for $\hat\lambda$ coincides with the ML
estimate of the noise precision. Since, as we will see later in the
section, only the first moment of $\lambda$ is needed to compute
other messages, it is sufficient to pass just this value to the rest
of the graph.

\subsection{Channel Weights Subgraph}

The channel weights subgraph includes the graphical description of
the factor $f_{\mathrm{C}}$ in \eqref{eq:syst_func}. We will present
in the following two alternative subgraphs representing two possible
definitions of $f_{\mathrm{C}}$: in the first one, coined
\emph{joint} channel weights subgraph, all channel weights for all
transmit antennas are grouped together in a single variable node
$\h$; in the second one, which we refer to as \emph{disjoint}
channel weights subgraph, the weights are split into $M$ variable
nodes $\h_1,\dots,\h_M$ each of them containing the channel weights
associated with an individual transmit antenna.

\subsubsection{Joint Channel Weights Model}\label{Sec:JCW}

\begin{figure}
\centering
\begin{psfrags}
\psfrag{fac}[c][c][1]{$f_{\mathrm{O}}$}
\psfrag{var}[c][c][1]{$\boldsymbol{h}$}
\psfrag{fac2}[c][c][1]{$f_{\mathrm{C}}$}
\psfrag{m1}[c][c][1]{$\boldsymbol{M}_{\mathrm{M}}$}
\psfrag{m2}[c][c][1]{$\boldsymbol{M}_{\mathrm{N}}$}
\psfrag{m3}[c][c][1]{$\boldsymbol{N}_{\mathrm{N}}$}
\psfrag{m4}[c][c][1]{$\boldsymbol{M}_{\mathrm{C}}$}
\psfrag{m5}[c][c][1]{$\boldsymbol{N}_{\mathrm{C}}$}
\psfrag{m6}[c][c][1]{$\boldsymbol{N}_{\mathrm{M}}$}
\psfrag{m9}[c][c][1]{$m_{f_{\mathrm{O}}\to \boldsymbol{h}}$}
\psfrag{m7}[c][c][1]{$n_{\boldsymbol{h}\to f_{\mathrm{O}} }$}
\psfrag{m8}[c][c][1]{$m_{f_{\mathrm{C}}\to \boldsymbol{h} }$}
\psfrag{cloud3}[c][c][1]{Joint Channel Weights}
\includegraphics[width=0.8\textwidth]{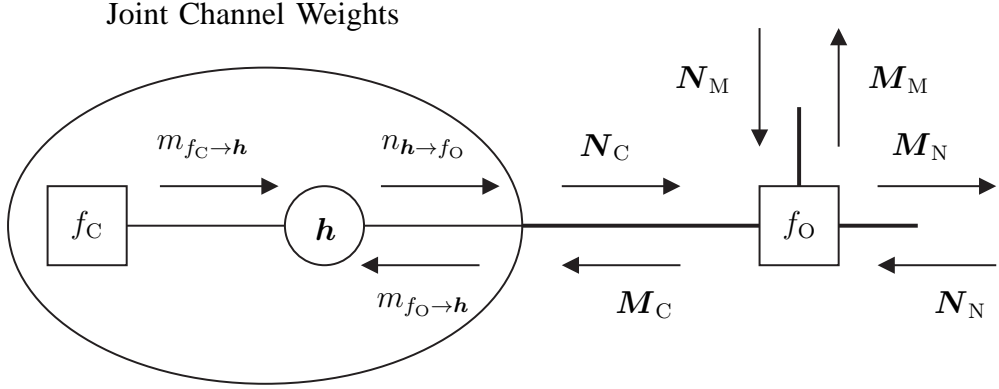}
\end{psfrags}
\caption{Subgraph corresponding to the prior model of the joint
channel weights.}\label{fig:FacGraphJCE}
\end{figure}

The joint channel weights subgraph is obtained from the following
definition:
\begin{equation*}
f_{\mathrm{C}}(\h)\triangleq p(\h)
\end{equation*}
with $p(\h)$ denoting the prior pdf of the vector of channel weights
$\h$. Using this model for $f_{\mathrm{C}}$ leads to defining the
factor and variable node sets as
\begin{align*}
\mathcal{A}_{\mathrm{C}}= & \{f_{\mathrm{C}}\}\\
\mathcal{I}_{\mathrm{C}}= & \{\h\}.
\end{align*}
The factor graph describing the joint channel weight option is
presented in Fig.~\ref{fig:FacGraphJCE}. As there is only one
variable node connected to the factor node $f_{\mathrm{O}}$, the set
of input messages to the channel weights subgraph is simply
$\Mess{C}=\{\mess{f_{\mathrm{O}}}{\h}\}$ and the set of output
messages is the singleton $\Ness{C}=\{\ness{\h}{f_{\mathrm{O}}}\}$.

The message from $f_{\mathrm{O}}$ to $\h$ is given by
\begin{equation*}
\mess{f_{\mathrm{O}}}{\h}(\h) = \exp\{\langle \log
f_{\mathrm{O}}(\y,\x,\h,\lambda)\rangle_{\Ness{M}\Ness{N}}\} \propto
\exp\left\{ -\hat\lambda\left( \|\y-\hat\X\h\|^2 + \h\herm\D\h
\right) \right\}
\end{equation*}
with $\D=\langle \X\herm\X\rangle_{\Ness{M}}-\hat\X\herm\hat\X$.
Hence, $\mess{f_{\mathrm{O}}}{\h}(\h)$ is proportional to a Gaussian
pdf. We also impose the prior $p(\h)$ to be Gaussian, which yields
the message
\begin{equation*}
\mess{f_{\mathrm{C}}}{\h}(\h) = p(\h) \propto \exp\left\{ -(\h -
\h_{\textrm{prior}})\herm\covMat{\h_{\textrm{prior}}}^{-1}(\h -
\h_{\textrm{prior}})\right\}.
\end{equation*}
For most practical channels it is reasonable to assume that
$\h_{\textrm{prior}}=0$. The receiver needs an estimate of the prior
covariance of the channel $\covMat{\h_{\textrm{prior}}}$. In order
to obtain the outgoing message $\ness{\h}{f_{\mathrm{O}}}(\h)$, the
two incoming messages are combined, leading to
\begin{equation*}
\ness{\h}{f_{\mathrm{O}}}(\h)\propto\mess{f_{\mathrm{O}}}{\h}(\h)\mess{f_{\mathrm{C}}}{\h}(\h)
\propto \exp\left\{-(\h-\hat\h)\herm\covEst{\h}^{-1}(\h-\hat\h).
\right\}
\end{equation*}
Thus, $\ness{\h}{f_{\mathrm{O}}}$ is proportional to a Gaussian pdf
with covariance matrix
\begin{equation*}
\covEst{\h}=\left(\hat\lambda\hat\X\herm\hat\X+\hat\lambda\D+\covMat{\h_{\textrm{prior}}}^{-1}\right)^{-1}
\end{equation*}
and mean value
\begin{equation*}
\hat\h=\covEst{\h}\left(\hat\lambda\hat\X\herm\y+\covMat{\h_{\textrm{prior}}}^{-1}\h_{\textrm{prior}}\right).
\end{equation*}

\subsubsection{Disjoint Channel Weights Model}\label{Sec:DCW}

\begin{figure}
\centering
\begin{psfrags}
\psfrag{fac}[c][c][1]{$f_{\mathrm{O}}$}
\psfrag{var2}[c][c][1]{$\boldsymbol{h}_1$}
\psfrag{fac2}[c][c][1]{$f_{\mathrm{C}_1}$}
\psfrag{var3}[c][c][1]{$\boldsymbol{h}_M$}
\psfrag{fac3}[c][c][1]{$f_{\mathrm{C}_M}$}
\psfrag{m1}[c][c][1]{$\boldsymbol{M}_{\mathrm{M}}$}
\psfrag{m2}[c][c][1]{$\boldsymbol{M}_{\mathrm{N}}$}
\psfrag{m3}[c][c][1]{$\boldsymbol{N}_{\mathrm{N}}$}
\psfrag{m4}[c][c][1]{$\boldsymbol{M}_{\mathrm{C}}$}
\psfrag{m5}[c][c][1]{$\boldsymbol{N}_{\mathrm{C}}$}
\psfrag{m6}[c][c][1]{$\boldsymbol{N}_{\mathrm{M}}$}
\psfrag{m7}[c][c][1]{$n_{\boldsymbol{h}_1\to f_{\mathrm{O}} }$}
\psfrag{m8}[c][c][1]{$m_{f_{\mathrm{C}_1}\to \boldsymbol{h}_1 }$}
\psfrag{m9}[c][c][1]{$m_{f_{\mathrm{O}}\to \boldsymbol{h}_1}$}
\psfrag{me1}[c][c][1]{$n_{\boldsymbol{h}_M\to f_{\mathrm{O}} }$}
\psfrag{me2}[c][c][1]{$m_{f_{\mathrm{C}_M}\to \boldsymbol{h}_M }$}
\psfrag{me3}[c][c][1]{$m_{f_{\mathrm{O}}\to \boldsymbol{h}_M}$}
\psfrag{cloud3}[c][c][1]{Disjoint Channel Weights}
\includegraphics[width=0.8\textwidth]{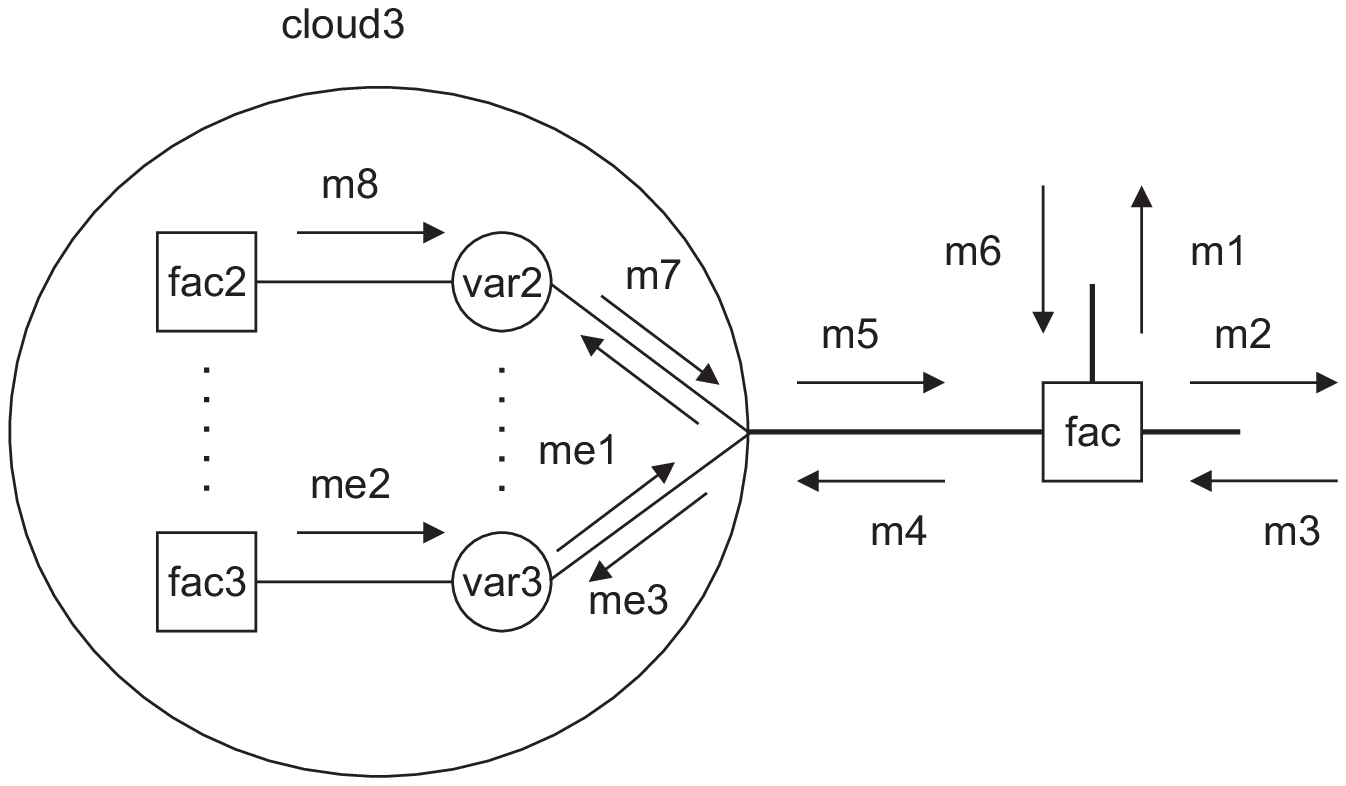}
\end{psfrags}
\caption{Subgraph corresponding to the prior model of the disjoint
channel weights.}\label{fig:FacGraphSCE}
\end{figure}

The disjoint channel weights subgraph is obtained by factorizing
$f_{\mathrm{C}}$ with respect to each transmitter. More
specifically, we define
\begin{equation*}
f_{\mathrm{C}}(\h)=\prod_{m=1}^{M}f_{\mathrm{C}_m}(\h_m)
\end{equation*}
with $f_{\mathrm{C}_m}(\h_m)\triangleq p(\h_m)$, $m=1,\dots,M$
denoting the prior pdf of the channel weights for the $m$th transmit
antenna. We also specify the sets
\begin{align*}
\mathcal{A}_{\mathrm{C}}= & \{f_{\mathrm{C}_m}|m=1,\dots,M\}\\
\mathcal{I}_{\mathrm{C}}= & \{\h_m|m=1,\dots,M\}.
\end{align*}
Fig.~\ref{fig:FacGraphSCE} shows the factor graph of the disjoint
channel weights model with the above definitions. With this
configuration, the channel weight vector $\h$ is split into $M$
variable nodes $\h_1,\dots,\h_M$, each of them containing the
weights associated with one transmit antenna. Each of these variable
nodes is furthermore connected to a factor node $f_{\mathrm{C}_m}$.
Due to this separation, the set of incoming messages reads
$\Mess{C}=\left\{\mess{f_{\mathrm{O}}}{\h_m}|m=1,\dots,M\right\}$,
while the set of outgoing messages is
$\Ness{C}=\left\{\ness{\h_m}{f_{\mathrm{O}}}|m=1,\dots,M\right\}$.
With this structure, the channel weight vectors are estimated
sequentially by iterating through the transmit antenna index $m$.

For the $m$th transmit antenna, the incoming message reads
\begin{align*}
\mess{f_{\mathrm{O}}}{\h_m}(\h_m)= & \exp\left\{ \langle\log
f_{\mathrm{O}}(\y,\x,\h,\lambda)\rangle_{\Ness{M}\Ness{N}\Ness{C}^{(m)}} \right\}\\
\propto & \exp\left\{-\hat\lambda\left(\big\|\y-\sum_{m^{\prime}\neq
m}\hat\X_{m^{\prime}}\hat\h_{m^{\prime}}-\hat\X_m\h_m\big\|^2+\h_m\herm\D_m\h_m
\right)\right\}
\end{align*}
where $\Ness{C}^{(m)}=\left\{\ness{\h_{m^{\prime}
}}{f_{\mathrm{O}}}\right\}_{\substack{\forall m^{\prime} =1,\dots,M \\
m^{\prime} \neq m}}$ denotes the set of all output channel weight
messages except the $m$th one. Furthermore,
$\hat\h_{m^{\prime}}=\langle
\h_{m^{\prime}}\rangle_{\Ness{C}^{(m)}}$, $\hat\X_m=\langle
\X_m\rangle_{\Ness{M}}$ and $\D_m=\langle
\X_m\herm\X_m\rangle_{\Ness{M}}-\hat\X_m\herm\hat\X_m$. Again,
$\mess{f_{\mathrm{O}}}{\h_m}$ is observed to be proportional to a
Gaussian pdf. Analogously to the joint channel weights case, we need
to specify the prior of each individual channel vector $\h_m$.
Defining them as Gaussians leads to the message
\begin{equation*}
\mess{f_{\mathrm{C}_m}}{\h_m}(\h_m)=p(\h_m)\propto \exp\left\{
-(\h_m-\h_{m,\textrm{prior}})\herm\covMat{\h_{m,\textrm{prior}}}^{-1}(\h_m-\h_{m,\textrm{prior}})\right\}
\end{equation*}
where, once more, the receiver requires estimates of the prior
parameters of the channel for each transmitter. The outgoing message
from variable node $\h_m$ is obtained by multiplying both incoming
messages, leading to
\begin{equation*}
\ness{\h_m}{f_{\mathrm{O}}}(\h_m)\propto\mess{f_{\mathrm{O}}}{\h_m}(\h_m)\mess{f_{\mathrm{C}_m}}{\h_m}(\h_m)\propto\exp\left\{
-(\h_m-\hat\h_{m})\herm\covEst{\h_{m}}^{-1}(\h_m-\hat\h_{m})\right\},
\end{equation*}
which equals, up to a proportionality constant, a Gaussian pdf with
covariance matrix
\begin{equation*}
\covEst{\h_{m}} =
\left(\hat\lambda\hat\X_m\herm\hat\X_m+\hat\lambda\D_m+\covMat{\h_{m,\textrm{prior}}}^{-1}\right)^{-1}
\end{equation*}
and mean value
\begin{equation*}
\hat\h_{m}=\covEst{\h_{m}}\left(\hat\lambda\hat\X_m\herm\left(\y-\sum_{m^{\prime}\neq
m}\hat\X_{m^{\prime}}\hat\h_{m^{\prime}}\right)+\covMat{\h_{m,\textrm{prior}}}^{-1}\h_{m,\textrm{prior}}\right).
\end{equation*}
It is important to note that every time a new message
$\ness{\h_m}{f_{\mathrm{O}}}$ is computed, the set of messages
$\Mess{C}$ needs to be recomputed again, as all
$\mess{f_{\mathrm{O}}}{\h_{m^{\prime}}}, m^{\prime}\neq m$ depend on
the updated messages $\ness{\h_m}{f_{\mathrm{O}}}$.

\subsection{Modulation and Coding Subgraph}

The modulation and coding subgraph describes the factor
$f_{\mathrm{M}}$ in \eqref{eq:syst_func}. We choose to factorize
this factor according to
\begin{align*}
& f_{\mathrm{M}}(\x,\cvec,\bvec)= \\ &
\prod_{m=1}^{M}f_{\mathcal{P}_m}(\x_m^{(p)})f_{\mathcal{M}_m}(\x_m^{(d)},c_{m,1},\dots,c_{m,C_m})
f_{\mathcal{C}_m}(c_{m,1},\dots,c_{m,C_m},u_{m,1},\dots,
u_{m,U_m})\prod_{i=1}^{U_m}f_{u_{m,i}}(u_{m,i})
\end{align*}
where $f_{\mathcal{P}_m}(\x_m^{(p)})\triangleq p(\x_m^{(p)})$
denotes the prior pdf of the pilot symbols transmitted from the
$m$th transmitter,
$f_{\mathcal{M}_m}(\x_m^{(d)},c_{m,1},\dots,c_{m,C_m})\triangleq
p(\x_m^{(d)}|c_{m,1},\dots,c_{m,C_m})$ denotes the modulation
constraints on the data symbols of the $m$th transmitter,
$f_{\mathcal{C}_m}(c_{m,1},\dots,c_{m,C_m},u_{m,1},\dots,
u_{m,U_m})\triangleq p(c_{m,1},\dots,c_{m,C_m}|u_{m,1},\dots,
u_{m,U_m})$ represents the code constraints for the $m$th codeword
and $f_{u_{m,i}}(u_{m,i})\triangleq p(u_{m,i})$ is the prior pdf of
the $i$th information bit transmitted by the $m$th antenna. In
addition, the vectors $\x_m^{(p)}$ and $\x_m^{(d)}$ contain,
respectively, the modulated pilot and data symbols transmitted from
the $m$th antenna. Finally, $C_m$ and $U_m$ denote the number of
coded and information bits respectively transmitted in a codeword
from the $m$th antenna. Using this factorization of
$f_{\mathrm{M}}$, we define the sets $\mathcal{A}_{\mathrm{M}}$ and
$\mathcal{I}_{\mathrm{M}}$ as
\begin{align*}
\mathcal{A}_{\mathrm{M}}= & \{f_{\mathcal{P}_m}|m=1,\dots,M\}\cup
\{f_{\mathcal{M}_m}|m=1,\dots,M\}\cup
\{f_{\mathcal{C}_m}|m=1,\dots,M\} \\ & \cup
\{f_{u_{m,i}}|m=1,\dots,M,
i=1\dots U_m\} \\
\mathcal{I}_{\mathrm{M}}= & \{\x_m^{(p)}|m=1\dots,M\}\cup
\{\x_m^{(d)}|m=1\dots,M\} \cup \{c_{m,i}|m=1,\dots,M, i=1\dots C_m\}
\\ & \cup \{u_{m,i}|m=1,\dots,M, i=1\dots U_m\}.
\end{align*}

\begin{figure}
\centering
\begin{psfrags}\small
\psfrag{fac1}[c][c][1]{$f_{\mathrm{O}}$}
\psfrag{m1}[c][c][1]{$\boldsymbol{M}_{\mathrm{M}}$}
\psfrag{m2}[c][c][1]{$\boldsymbol{M}_{\mathrm{N}}$}
\psfrag{m3}[c][c][1]{$\boldsymbol{N}_{\mathrm{N}}$}
\psfrag{m4}[c][c][1]{$\boldsymbol{M}_{\mathrm{C}}$}
\psfrag{m5}[c][c][1]{$\boldsymbol{N}_{\mathrm{C}}$}
\psfrag{m6}[c][c][1]{$\boldsymbol{N}_{\mathrm{M}}$}
\psfrag{var1}[c][c][1]{$\boldsymbol{x}_1^{(p)}$}
\psfrag{var2}[c][c][1]{$\boldsymbol{x}_M^{(p)}$}
\psfrag{var3}[c][c][1]{$\boldsymbol{x}_1^{(d)}$}
\psfrag{var4}[c][c][1]{$\boldsymbol{x}_M^{(d)}$}
\psfrag{var5}[c][c][1]{$c_{1,1}$}
\psfrag{var6}[c][c][1]{$c_{1,C_1}$}
\psfrag{var7}[c][c][1]{$u_{1,1}$}
\psfrag{var8}[c][c][1]{$u_{1,U_1}$}
\psfrag{fac2}[c][c][1]{$f_{\mathcal{P}_1}$}
\psfrag{fac3}[c][c][1]{$f_{\mathcal{P}_M}$}
\psfrag{fac4}[c][c][1]{$f_{{\cal M}_1}$}
\psfrag{fac5}[c][c][1]{$f_{{\cal M}_M}$}
\psfrag{fac6}[c][c][1]{$f_{u_{1,1}}$}
\psfrag{fac7}[c][c][1]{$f_{u_{1,U_1}}$}
\psfrag{n1}[c][c][1]{$\ness{\boldsymbol{x}_M^{(p)}}{f_{\mathrm{O}}}$}
\psfrag{n3}[c][c][1]{$\ness{\boldsymbol{x}_1^{(d)}}{f_{\mathrm{O}}}$}
\psfrag{n4}[c][c][1]{$\mess{f_{\mathrm{O}}}{\boldsymbol{x}_1^{(d)}}$}
\psfrag{n5}[c][c][1]{$\mess{f_{\mathcal{P}_M}}{\boldsymbol{x}_M^{(p)}}$}
\psfrag{n7}[c][c][1]{$\mess{f_{{\cal
M}_1}}{\boldsymbol{x}_1^{(d)}}$}
\psfrag{n8}[c][c][1]{$\ness{\boldsymbol{x}_1^{(d)}}{f_{{\cal
M}_1}}$} \psfrag{cloud}[c][c][1]{Modulation and coding}
\psfrag{box}[c][c][1]{$f_{{\cal C}_1}$}
\includegraphics[width=0.95\textwidth]{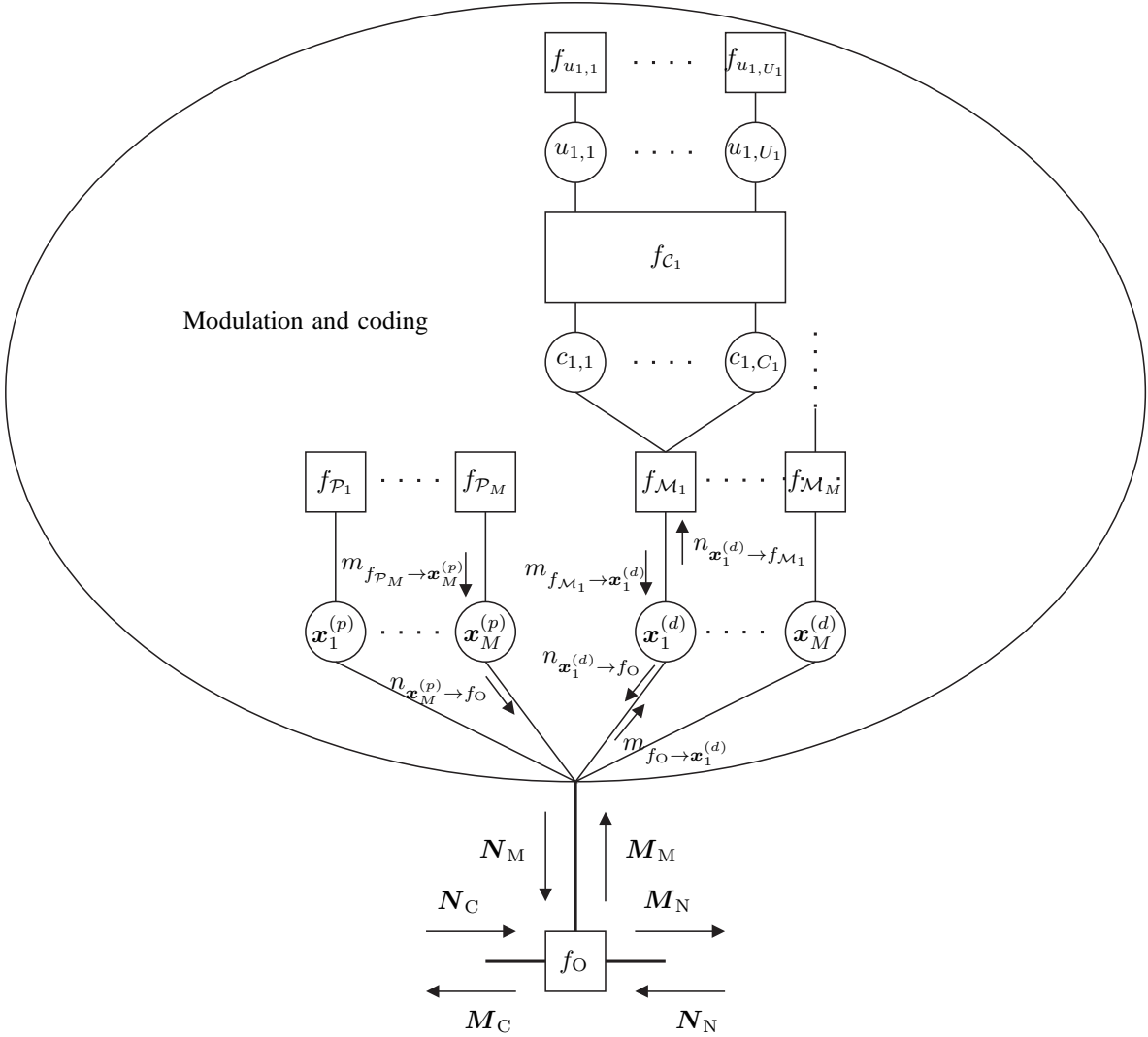}
\end{psfrags}
\caption{Subgraph corresponding to the modulation and coding
constraints.} \label{fig:FacGraphDet}
\end{figure}

The factor graph with the modulation and coding constraints is shown
in Fig.~\ref{fig:FacGraphDet}. As it can be observed, the modulated
symbols have been separated into different variable nodes according
to the transmit antenna index $m$ from which they are sent. The
symbols corresponding to each transmit antenna port have been
further subdivided into two different variable nodes $\x_m^{(p)}$
and $\x_m^{(d)}$, the first containing the pilot symbols and the
second containing the modulated data symbols. The modulated data
symbols $\x_m^{(d)}$ are connected to the encoded bits
$c_{m,1},\dots,c_{m,C_m}$ via the modulation factor node $f_{{\cal
M}_m}$, which describes the mapping of bits onto a complex
constellation. The coded bits are, in turn, related to the
information bits $u_{m,1},\dots,u_{m,U_m}$ through the specific
channel code and interleaving scheme utilized, which is represented
in a simplified manner by the factor $f_{\mathcal{C}_m}$ in
Fig.~\ref{fig:FacGraphDet}. Finally, every information bit $u_{m,i}$
has an associated prior probability represented by the factor node
$f_{u_{m,i}}$. For the vast majority of applications, however, the
values of the bits will be assumed to be equiprobable. With the
proposed structure, the set of incoming messages is defined as
$\Mess{M}=\left\{\mess{f_{\mathrm{O}}}{\x_m^{(p)}}|m=1,\dots,M\right\}\cup
\left\{\mess{f_{\mathrm{O}}}{\x_m^{(d)}}|m=1,\dots,M\right\}$, while
the set of outgoing messages becomes
$\Ness{M}=\left\{\ness{\x_m^{(p)}}{f_{\mathrm{O}}}|m=1,\dots,M\right\}\cup
\left\{\ness{\x_m^{(d)}}{f_{\mathrm{O}}}|m=1,\dots,M\right\}$.

In order to ease the derivation of the messages for this subgraph,
we can re-write $f_{\mathrm{O}}(\y,\x,\h,\lambda)$ as
\begin{align*}
f_{\mathrm{O}}(\y,\x,\h,\lambda) & \propto
\lambda^{KNL}\exp\left\{-\lambda\Big\|\y^{(d)} -
\sum_{m=1}^M\Hmat_m^{(d)}\x_m^{(d)}\Big\|^2-\lambda\Big\|\y^{(p)} -
\sum_{m=1}^M\Hmat_m^{(p)}\x_m^{(p)}\Big\|^2\right\}
\end{align*}
where the contribution of pilot and data symbols has been split into
two separate terms. We start by computing the message that factor
node $f_{\mathrm{O}}$ sends to $\x_{m}^{(d)}$:
\begin{align}
\mess{f_{\mathrm{O}}}{\x_{m}^{(d)}}(\x_m^{(d)}) & = \exp\left\{
\langle\log
f_{\mathrm{O}}(\y,\x,\h,\lambda)\rangle_{\Ness{N}\Ness{C}\Ness{M}^{(m)}} \right\}\nonumber\\
& \propto
\exp\left\{-\hat\lambda\left(\Big\|\y^{(d)}-\sum_{m^{\prime}\neq
m}\hat\Hmat_{m^{\prime}}^{(d)}\hat\x_{m^{\prime}}^{(d)}-\hat\Hmat_{m}^{(d)}\x_{m}^{(d)}\Big\|^2+(\x_m^{(d)})\herm\C_m^{(d)}\x_m^{(d)} \right.\right.\nonumber\\
&+\left.\left.\sum_{m^{\prime}\neq m}\left(
(\x_m^{(d)})\herm\C_{mm^{\prime}}^{(d)}\hat\x_{m^{\prime}}^{(d)}+(\hat\x_{m^{\prime}}^{(d)})\herm(\C_{mm^{\prime}}^{(d)})\herm\x_m^{(d)}\right)\right)\right\}\label{eq:mess_x_vmp}.
\end{align}
In the above expression, and similarly to previous definitions,
$\hat\x_{m^{\prime}}^{(d)}=\langle\x_{m^{\prime}}^{(d)}\rangle_{\Ness{M}}$,
$\hat\Hmat_{m^{\prime}}^{(d)}=\langle\Hmat_{m^{\prime}}^{(d)}\rangle_{\Ness{C}}$,
$\C_m^{(d)}=\langle(\Hmat_{m}^{(d)})\herm\Hmat_{m}^{(d)}\rangle_{\Ness{C}}-(\hat\Hmat_{m}^{(d)})\herm\hat\Hmat_{m}^{(d)}$
and
$\C_{mm^{\prime}}^{(d)}=\langle(\Hmat_{m}^{(d)})\herm\Hmat_{m^{\prime}}^{(d)}\rangle_{\Ness{C}}-(\hat\Hmat_{m}^{(d)})\herm\hat\Hmat_{m^{\prime}}^{(d)}$.
Additionally,
$\Ness{M}^{(m)}=\{\ness{\x_i^{(p)}}{f_{\mathrm{O}}}|i=1,\dots,M\}\cup
\{\ness{\x_i^{(d)}}{f_{\mathrm{O}}}|i=1,\dots,M,i\neq m\}$ denotes
the set of all outgoing detection messages except
$\ness{\x_m^{(d)}}{f_{\mathrm{O}}}$. The message in
\eqref{eq:mess_x_vmp} is proportional to a Gaussian pdf with
covariance matrix
\begin{equation*}
\covEst{\x_{m,\textrm{VMP}}^{(d)}}=\hat\lambda^{-1}\left((\hat\Hmat_m^{(d)})\herm\hat\Hmat_m^{(d)}+\C_m^{(d)}\right)^{-1}
\end{equation*}
and mean
\begin{equation*}
\hat\x_{m,\textrm{VMP}}^{(d)}=\hat\lambda\covEst{\x_{m,\textrm{VMP}}^{(d)}}\left((\hat\Hmat_m^{(d)})\herm\left(\y^{(d)}-\sum_{m^{\prime}\neq
m}\hat\Hmat_{m^{\prime}}^{(d)}\hat\x_{m^{\prime}}^{(d)}\right)-\sum_{m^{\prime}\neq
m}\C_{mm^{\prime}}^{(d)}\hat\x_{m^{\prime}}^{(d)}\right).
\end{equation*}
The outgoing message $\ness{\x_m^{(d)}}{f_{\mathrm{O}}}(\x_m^{(d)})$
is obtained by multiplying the messages
$\mess{f_{\mathrm{O}}}{\x_{m}^{(d)}}(\x_m^{(d)})$ and
$\mess{f_{{\cal M}_m}}{x_m^{(d)}}$. In this case, $\mess{f_{{\cal
M}_m}}{x_m^{(d)}}$ is a SP message reading
\begin{equation}\label{eq:mess_M_x}
\mess{f_{{\cal
M}_m}}{x_m^{(d)}}\propto\prod_{i=1}^{N_d}\left(\sum_{s\in{\cal
S}_m}\beta_{x_m^{(d)}(i)}(s)\delta(x_m^{(d)}(i)-s)\right)
\end{equation}
where ${\cal S}_m$ is the modulation set for user $m$ and
$\beta_{x_m^{(d)}(i)}(s)$ represents the extrinsic values of
$x_m^{(d)}(i)$ for each constellation point $s\in{\cal S}_m$,
obtained from the SP demodulator and decoder. The combined message
fed back to the observation factor node reads
\begin{align}
\ness{\x_m^{(d)}}{f_{\mathrm{O}}}(\x_m^{(d)})& \propto
\mess{f_{\mathrm{O}}}{\x_{m}^{(d)}}(\x_m^{(d)})\mess{f_{{\cal
M}_m}}{x_m^{(d)}}(\x_m^{(d)})\nonumber\\
& \propto \prod_{i=1}^{N_d}\left(\sum_{s\in{\cal
S}_m}\beta_{x_m^{(d)}(i)}(s)\exp\left\{\frac{-|s-\hat
x_{m,\textrm{VMP}}^{(d)}(i)|^2}{\sigma_{x_m^{(d)}}^2(i)}\right\}\delta(x_m^{(d)}(i)-s)\right),\label{eq:mess_x_y}
\end{align}
where $\sigma_{x_m^{(d)}}^2(i)$ is the $i$th entry in the main
diagonal of $\covEst{\x_{m,\textrm{VMP}}^{(d)}}$. It can be observed
that the message factorizes with respect to the individual modulated
symbols ${x_m^{(d)}(i)}$, so the mean and variance of each data
symbol can be computed independently and used to build the mean
vector $\hat\x$ and the covariance matrix $\covEst{\x}$ by inserting
the updated mean and variances in their corresponding positions.

It is important to note that, because the factor node $f_{{\cal
M}_m}$ is a SP factor node, the message $\ness{\x_m^{(d)}}{f_{{\cal
M}_m}}$ is obtained by multiplying all messages received at variable
node $\x_m^{(d)}$ except the message coming from $f_{{\cal M}_m}$,
which in this particular setup reduces to
\begin{equation*}
\ness{\x_m^{(d)}}{f_{{\cal M}_m}}(\x_m^{(d)}) =
\mess{f_{\mathrm{O}}}{\x_{m}^{(d)}}(\x_m^{(d)}).
\end{equation*}
All message-passing among the modulation factor nodes, coded bits
and information bits is completed by using the standard SP
algorithm, and will therefore not be described here.

It remains to describe the income and outcome messages involving
pilot symbols. As pilot symbols are known by the receiver, their
prior distribution is $p(x_m^{(p)}(i))=\delta(x_m^{(p)}(i)-p_m(i))$
with $p_m(i)$ denoting the $i$th pilot symbol sent from transmit
antenna $m$. This imposes that the outgoing message
$\ness{\x_m^{(p)}}{f_{\mathrm{O}}}$ is also a Dirac delta, which can
also be described as a degenerate Gaussian message with mean
$\hat\x_m^{(p)}=\boldsymbol{p}_m$ and covariance
$\covEst{\x_m^{(p)}}=\boldsymbol{0}$.


%

%
%

\section{Simulation Results}
\label{Sec:Results}
In this section, we propose a number of receiver structures based on
the derivations made in Section~\ref{Sec:GenRec} and evaluate their
performance by means of Monte-Carlo simulations. First, we present
the parameters of the MIMO-OFDM system considered, followed by a
description of the specific receiver structures that will be
evaluated. Finally, the BER performance results obtained are
presented and discussed.

\subsection{Description of the MIMO-OFDM System}

We begin by describing the MIMO-OFDM system used for obtaining the
numerical results. Its main parameters are summarized in
Table~\ref{Tab:param}. We consider an OFDM system with $M=N=2$
antennas at both transmitter and receiver ends. Two streams of
random bits are independently encoded using a convolutional code
with rate 1/3 and generating polynomials 133, 171 and 165 (octal).
After channel interleaving, the coded bits are mapped onto symbols
of a QPSK or 16QAM constellation (with Gray mapping) which are then
inserted into an OFDM frame consisting of $L=7$ OFDM symbols with
$K=75$ subcarriers and with a subcarrier spacing of 15kHz. Part of
the time-frequency elements are reserved for the transmission of
pilot symbols. We specify the following pilot patterns: pilot
symbols are transmitted in the first and fifth OFDM symbol of the
frame, with a frequency spacing of 12 subcarriers, resulting in a
total of 13 pilot symbols per frame. Note that both transmit
antennas share the same time-frequency elements for the simultaneous
transmission of pilot symbols. Pilot symbols are randomly chosen
from a QPSK constellation.

Realizations of the channel time-frequency response are randomly
generated using the extended typical urban (ETU) model from the 3GPP
LTE standard~\cite{3GPP36104} with 9 Rayleigh-fading taps. The
channel responses corresponding to two different transmitters are
uncorrelated and remain static over the duration of an OFDM frame. A
new channel response is generated for each OFDM frame, with the
responses of two different frames being also uncorrelated.

\begin{table}
\centering \caption{Parameters of the simulated OFDM system}
\label{Tab:param}
\begin{tabular}{c|c}
Parameter & Value \\ \hline
Tx antennas ($M$) & 2 \\
Rx antennas ($N$) & 2 \\
Subcarriers ($K$) & 75 \\
OFDM symbols ($L$) & 7 \\
Subcarrier spacing ($\Delta f$) & 15 kHz\\
Channel coding & 1/3 Convolutional\\
Symbol mapping & 16QAM\\
Pilot symbols & 13\\
Channel model & 3GPP ETU
\end{tabular}

\end{table}

\subsection{Receiver Structures}\label{Sec:RecStr}

We introduce now the specific receiver architectures that will be
evaluated in this section. All receivers are based on the generic
message-passing receiver presented in Section~\ref{Sec:GenRec}. The
messages exchanged can be obtained by particularizing the generic
messages according to the specific receiver configuration, as it
will be detailed in the following. We evaluate three main types of
VMP-SP receiver, which are described next.

\subsubsection{I-DJC-DD and I-DSC-DD Receivers}

First, we introduce a full iterative receiver using exactly the
messages derived in Section~\ref{Sec:GenRec}. 
The receiver operates by iteratively updating the beliefs of the
channel weight vector, the data symbols and information bits and,
finally, the noise precision parameter.

Initialization of the beliefs of the channel weights and the
transmitted symbols is required. The initialization of the channel
weights is obtained from a pilot-based joint linear minimum
mean-squared-error (LMMSE) channel estimator. For the initialization
of the transmitted symbols, maximum-likelihood detection (MLD) is
used, followed by soft-in soft-out (SISO) BCJR decoding. The belief
of the transmitted data symbols is set to a Gaussian pdf with mean
and covariance values obtained from soft modulation of the
a-posteriori probabilities (APP) of the coded bits obtained from the
SISO BCJR decoder. An initial estimate of the noise precision is
obtained as in Section~\ref{Sec:NPE}. After the initialization, a
full iteration of the receiver consists of updating the beliefs of
the channel weight vectors (using either the joint channel weight
model in Fig.~\ref{fig:FacGraphJCE} or the disjoint channel weight
model in Fig.~\ref{fig:FacGraphSCE}), a message-passing run on the
modulation and coding subgraph (updating the beliefs of transmitted
symbols, coded bits and information bits) and, finally, an update of
the noise precision parameter. Note that the message-passing
operations done through the channel code factor node can be replaced
by SISO BCJR decoding. In this case, the SP messages
$\ness{c_{m,k}}{f_{\mathcal{M}_m}}$ can be identified to be the
extrinsic values of the coded bits output by the BCJR decoder.


We refer to the described architectures as Iterative - Data-aided
Joint Channel estimation - Data Decoding (I-DJC-DD) for the receiver
using the joint channel weights model and Iterative - Data-aided
Sequential Channel estimation - Data Decoding (I-DSC-DD) for the
receiver obtained using the disjoint channel weights model.

\subsubsection{DJC-DD and DSC-DD Receivers}

We introduce now a class of receivers which perform iterative
data-aided channel weights and noise precision estimation together
with equalization and demodulation of the transmitted symbols.
Compared to the receivers presented before, channel decoding is left
outside of the iterative process, and is performed only once at the
end after convergence of the algorithm. The receiver capitalizes on
just the knowledge of the complex modulation structure of the
transmitted signal to refine its channel estimates, and not on the
code structure. This receiver architecture is obtained by applying a
special scheduling to the message computation and exchange between
the subgraphs. Specifically, no messages are passed from variable
nodes $\x_m^{(d)}$ to factor nodes $f_{\mathcal{M}_m}$ until the
last iteration of the algorithm. Instead, after the messages
$\mess{f_{\mathrm{O}}}{\x_m^{(d)}}$ are computed, the updated
message $\ness{\x_m^{(d)}}{f_{\mathrm{O}}}$ is directly computed
using \eqref{eq:mess_x_y}. To this end, an initial value of the
messages $\mess{f_{{\cal M}_m}}{x_m^{(d)}}$ is needed. This can be
obtained by setting
\begin{equation*}
\beta_{x_m^{(d)}(i)}(s) = \frac{1}{|\mathcal{S}_m|}\quad \forall
m=1,\dots,M, i=1,\dots,N_d, s\in\mathcal{S}_m
\end{equation*}
in \eqref{eq:mess_M_x}. In the expression above, $|\mathcal{S}_m|$
denotes the cardinality of the set $\mathcal{S}_m$. Note that this
initialization corresponds to assuming that all modulated symbols in
the constellation are equally likely, which is a valid assumption
when the information bits are equiprobable and the channel code is
regular.


As for the previous receivers, an initialization of the beliefs of
the channel weight vector, noise precision and transmitted symbols
is required. The channel weight vectors are initialized as a
Gaussian pdf, with mean obtained from a pilot-based LMSSE channel
estimator and null covariance. Similarly, the beliefs of the
transmitted symbols are also set to a Gaussian pdf with mean and
covariance values obtained from a MIMO MLD (no BCJR decoding is
done, as opposed to the I-DJC-DD and I-DSC-DD receivers). An initial
estimate of the noise precision is then obtained following the
procedure in Section~\ref{Sec:NPE}. After the initialization, the
receiver operates by iteratively updating the beliefs of the channel
weights (either jointly as in Fig.~\ref{fig:FacGraphJCE}, or
sequentially as in Fig.~\ref{fig:FacGraphSCE}), the transmitted
symbols and noise precision parameter. After convergence of the
algorithm (or maximum number of iterations attained), the messages
$\ness{\x_m^{(d)}}{f_{{\cal M}_m}}$ are computed, and a round of
decoding based on the SP algorithm is performed, yielding the
beliefs of the information bits.



We refer to these receivers as Data-aided Joint Channel estimation -
Data Decoding (DJC-DD) for the receiver using the joint channel
weight prior model (Section~\ref{Sec:JCW}) and Data-aided Sequential
Channel estimation - Data Decoding (DSC-DD) for the receiver using
the disjoint channel weight prior model (Section~\ref{Sec:DCW}).

\subsubsection{PSC-DD Receiver}

Finally we present a simple receiver consisting of a pilot-aided
channel estimator, a MIMO maximum likelihood detector (MLD) and data
decoding. The channel estimation module is based on the VMP-SP
generic receiver described in Section~\ref{Sec:GenRec}. It updates
iteratively the beliefs of the channel weight vectors corresponding
to each transmit antenna and the noise precision. To this end, the
channel estimator only exploits the pilot signals transmitted from
each transmit antenna and does not capitalize on data symbols to
refine its estimates.

In order to obtain this pilot-aided channel estimator from the
generic receiver architecture in Section~\ref{Sec:GenRec}, the
messages $\ness{\x_m^{(d)}}{f_{\mathrm{O}}}$ must be set to
\begin{equation*}
\ness{\x_m^{(d)}}{f_{\mathrm{O}}}(\x_m^{(d)}) =
\prod_i\delta(x_m^{(d)}(i)).
\end{equation*}
This enforces that data symbols are not employed for channel weight
estimation. In addition, the disjoint channel weights setup (see
Fig.~\ref{fig:FacGraphSCE}) is selected. With this configuration,
the output messages $\Ness{M}$ are constant, reflecting the
receiver's knowledge on the value of the pilot symbols. Hence,
expectations taken over $\Ness{M}$ in the channel weights and noise
precision subgraphs reduce to the value of the pilot symbols (or
zero for data symbols), with all second-order terms vanishing. Note
that, for this channel estimator, no update of the beliefs of the
data symbols is performed. Equalization and decoding are done
outside the VMP-SP framework.

Additionally, a small modification in the processing corresponding
to the noise precision subgraph is required. Note that, for the
computation of the message $\mess{f_{\mathrm{O}}}{\lambda}$, the
signal received at all --pilot and data-- subcarriers is used, while
only the signals received at pilot positions are utilized for
channel weight estimation. This can be avoided by restricting this
message to include only the observation at pilot positions, i.e.
calculating $\mess{f_{\mathrm{O}^{(p)}}}{\lambda}$ instead, where
\begin{align*}
f_{\mathrm{O}^{(p)}}(\y^{(p)},\x^{(p)},\h^{(p)},\lambda) \triangleq
p(\y^{(p)} | \x^{(p)},\h^{(p)},\lambda) & \propto
\lambda^{N^{(p)}}\exp\left\{-\lambda\left\|\y^{(p)} -
\X^{(p)}\h^{(p)}\right\|^2\right\},
\end{align*}
with $N^{(p)}$ denoting the total number of pilots in a frame.

The initialization for this estimator is simpler compared to that of
the other receivers. It consists of setting the beliefs of the
channel weight vector corresponding to each transmit antenna to a
Gaussian prior with zero mean and zero covariance, while an initial
value for the noise precision can be obtained from the signal
received at pilot subcarriers. The receiver operates by sequentially
updating the channel weight vectors corresponding to each
transmitter $\h_1,\dots,\h_M$ following the procedure described in
Section~\ref{Sec:DCW}. This is followed by an update of the noise
precision parameter. The channel responses belonging to each
transmit antenna obtained after convergence of the iterative
estimator are fed to a MIMO maximum likelihood detector (MLD),
followed by BCJR decoding. Thus, we can obtain BER performance
results and benchmark them with analogous receiver structures using
a different channel estimator.

As we will see in the performance evaluation, this iterative
estimator approximates the performance of a pilot-based joint LMMSE
channel estimator with perfect knowledge of the noise variance. The
iterative estimator, however, presents the advantage of avoiding
cumbersome matrix inversions depending on the specific values of the
pilot-symbols. This estimator was presented (outside the context of
message-passing techniques) in ~\cite{Navarro2009a}. A more detailed
discussion of the computational advantages of this estimator over
the LMMSE estimator is provided in this contribution.

In the following, we refer to this receiver as Pilot-aided
Sequential Channel estimation - Data Decoding (PSC-DD) receiver.

The main characteristics of all the receivers presented above are
summarized in Table~\ref{Tab:Receivers}.

\begin{table}\centering
\caption{Summary of receiver structures}\label{Tab:Receivers}
\begin{tabular}{c|c|c|c|c}
& \multicolumn{2}{|c|}{Initialization} &
\multicolumn{2}{|c}{Operation} \\ \hline Receiver  &  Channel
Weights &
Transmitted Symbols & Channel Weight Model & Demodulation \& Decoding \\
\hline PSC-DD & Null mean and covariance & -- & Disjoint & -- \\
DJC-DD & LMMSE estimator & ML detector & Joint & Demodulation only \\
DSC-DD & LMMSE estimator & ML detector & Disjoint & Demodulation
only
\\ I-DJC-DD & LMMSE estimator & MLD + BCJR & Joint & Demodulation and
decoding \\ I-DSC-DD & LMMSE estimator & MLD + BCJR & Disjoint &
Demodulation and decoding
\end{tabular}

\end{table}

\subsection{Numerical Results}


We evaluate separately the performance of the three architectures
described in Section~\ref{Sec:RecStr}, beginning with the simplest
scheme, the PSC-DD receiver; we follow with the DJC-DD and DSC-DD
receivers and conclude with the most advanced structures: the
I-DJC-DD and I-DSC-DD receivers.

\begin{figure}
\centering
\includegraphics[width=0.6\textwidth]{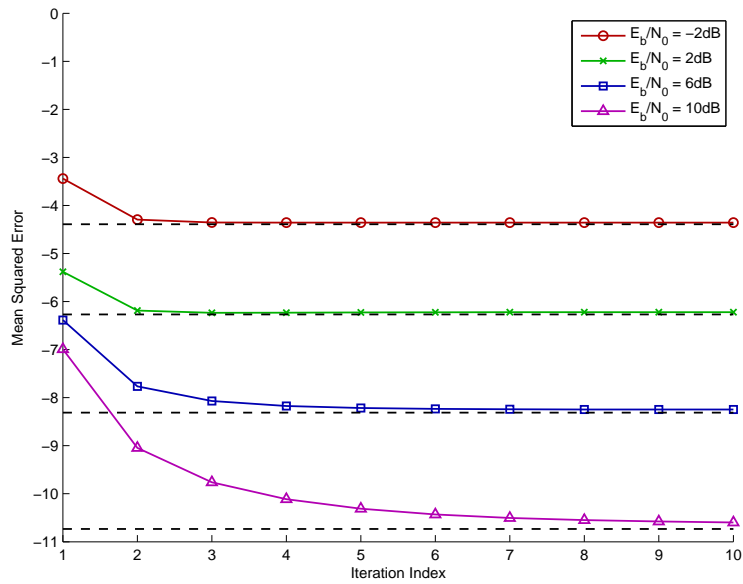}
\caption{MSE of the estimates of the channel weights for the PSC-DD
receiver versus the iteration index. 13 pilot symbols are inserted
per OFDM frame. The dashed black lines represent the MSE obtained
with pilot-based LMMSE joint channel estimation.}
\label{fig:MSEvsIt}
\end{figure}

\begin{figure}
\centering
\includegraphics[width=0.6\textwidth]{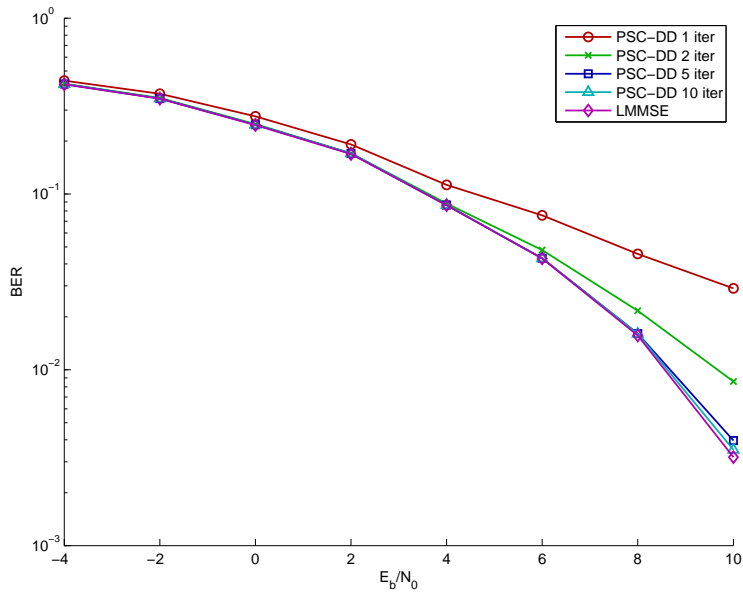}
\caption{BER as a function of $E_b/N_0$ for the PSC-DD receiver with
QPSK modulation. 13 pilot symbols are inserted per OFDM frame. The
BER performance of a similar receiver using LMMSE channel estimation
with knowledge of the noise variance is included as a reference.}
\label{fig:PbceBEer}
\end{figure}

In Fig.~\ref{fig:MSEvsIt}, the mean squared error (MSE) of the
estimates of the channel weights obtained with the PSC-DD receiver
is depicted. The MSE is plotted for three different $E_b/N_0$ values
as a function of the number of iterations performed. It is observed
that the performance of the sequential pilot-based estimator
approaches the performance of a joint LMMSE estimator with
sufficient number of iterations. It is especially interesting to
note the dependency of the number of iterations required for
convergence on the $E_b/N_0$ value. For $E_b/N_0=-2$dB and
$E_b/N_0=2$dB, between 2 and 3 iterations are sufficient to achieve
an MSE virtually equal to the LMMSE bound. When increasing $E_b/N_0$
to 6dB, however, a minimum number of 5 iterations is needed, and
about 10 iterations are required for $E_b/N_0=10$dB. Similar
observations can be made when evaluating the BER of the receiver
with QPSK modulation, as shown in Fig.~\ref{fig:PbceBEer}. Again,
the performance of the PSC-DD receiver equals that of the receiver
with the LMMSE estimator when enough iterations for the receiver to
converge have been run, and fewer iterations are needed the smaller
 $E_b/N_0$ is. These results suggest that the iterative channel
estimator in the PSC-DD receiver would be a good choice to obtain an
initial channel estimate for the more complex iterative structures
that we will discuss next. Furthermore, this channel estimator has
the additional benefit of outputting soft estimates (the beliefs) of
both the channel weights and the noise precision. Classical channel
estimators, on the other hand, typically require separate noise
estimation prior to the estimation of the channel weights, and only
provide hard (point) estimates.

\begin{figure}
\centering
\includegraphics[width=0.6\textwidth]{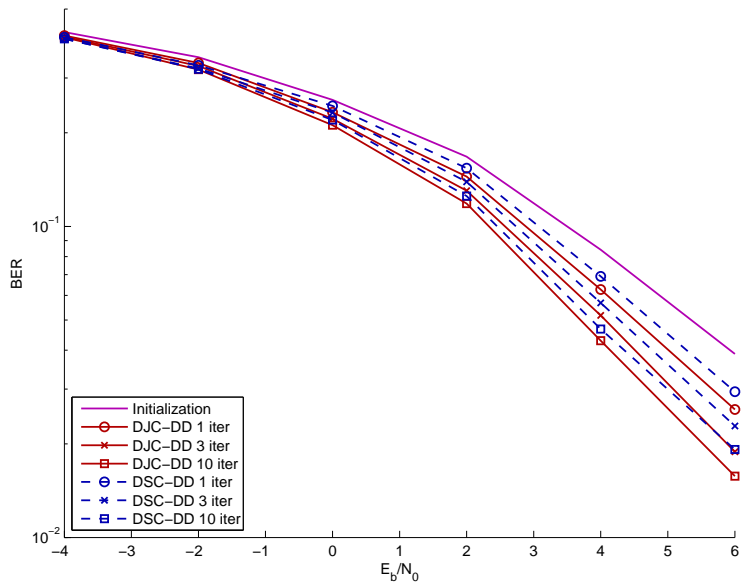}
\caption{BER as a function of $E_b/N_0$ for the DJC-DD and DSC-DD
receivers with QPSK modulation. 13 pilot symbols are inserted per
OFDM frame.} \label{fig:SLBer}
\end{figure}

BER results for the DJC-DD and DSC-DD receivers are portrayed in
Fig.~\ref{fig:SLBer}. The results have been obtained using a QPSK
constellation for the modulation of data symbols. They indicate that
a significant performance gain can be obtained by iteratively
updating the channel weights, transmitted data symbols and noise
precision parameter after the initialization, even though the
receiver does not capitalize on the code structure within the
iterative process. For both receivers (with joint and sequential
channel estimation), most of the improvement with respect to the
initialization is obtained in the first three iterations, with only
marginal gains obtained after further processing. Regarding the
channel estimation approach, DJC-DD leads to a slightly better
performance than DSC-DD in the whole simulated $E_b/N_0$ range; the
improved accuracy of the joint estimation approach comes at the
expense of a larger computational complexity, as it operates with
vectors and matrices of dimensions $M$ times as large as in the
sequential estimation approach, which can be a problem when
calculating the necessary matrix inversions.

Note that the receivers evaluated in Fig.~\ref{fig:SLBer} operate by
capitalizing on the structure of the constellation used for the
modulation of data symbols. Hence, their performance strongly
depends on the type of modulation used. Low-order modulations, such
as BPSK or QPSK, favor this receiver, as there is a relatively large
distance between the points in the constellation, allowing better
refining (through SP message-passing) of the VMP estimates of the
transmitted symbols. When using higher order modulations, however,
the receiver's performance suffers from the relatively small
distance between adjacent constellation points. Specifically for the
system investigated in this work, we found that  the DJC-DD and
DSC-DD receivers for 16-QAM or higher order modulations do not
improve the performance with respect to the initialization.

\begin{figure}
\centering
\includegraphics[width=0.6\textwidth]{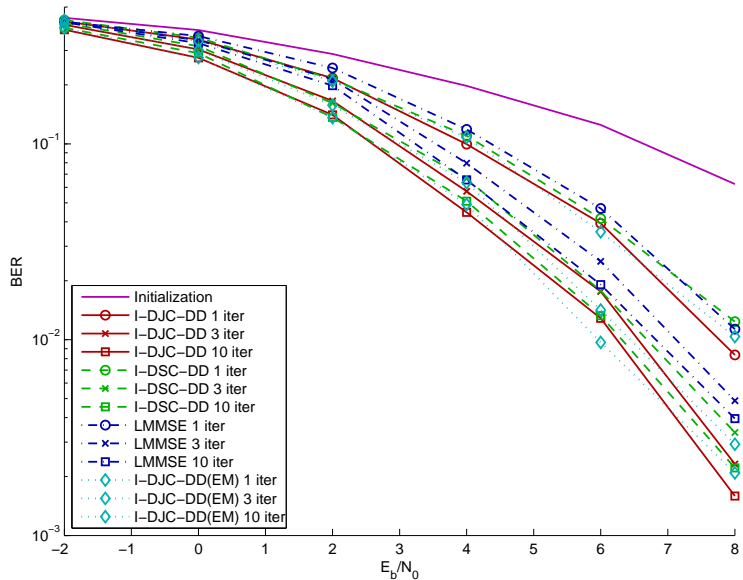}
\caption{BER as a function of $E_b/N_0$ for the I-DJC-DD and
I-DSC-DD receivers with 16-QAM modulation. 13 pilot symbols are
inserted per OFDM frame.} \label{fig:FRBer}
\end{figure}

The aforementioned problem with high-order modulations can be
circumvented with the inclusion of the channel code structure in the
iterative processing, as done in the I-DJC-DD and I-DSC-DD
receivers. The BER performance of both receivers with 16-QAM
modulated data symbols is depicted in Fig.~\ref{fig:FRBer}. For
benchmarking purposes, the BER performance of a heuristically
designed iterative receiver with analogous features to the I-DJC-DD
receiver is also plotted. We refer to this receiver as LMMSE
receiver, as the channel estimation and MIMO detection modules are
separately designed after the LMMSE principle. The LMMSE receiver is
based on the design proposed in~\cite{Wehinger2006} for a multiuser
CDMA receiver, and was adapted to MIMO-OFDM in~\cite{Kirkelund2010},
where a detailed description of its operational principles can be
found. In addition, the BER performance of a modified version of the
I-DJC-DD receiver has also been included. This receiver, which we
denote as I-DJC-DD(EM) receiver, results from applying the EM
restriction to the beliefs of the channel weights $\h$ and the noise
precision parameter $\lambda$. Thus, this receiver is identical to
the I-DJC-DD receiver except that the messages
$\ness{\h}{f_{\mathrm{O}}}$ and $\ness{\lambda}{f_{\mathrm{O}}}$ are
computed according to \eqref{eq:EM_mes}. This modified messages
imply that all terms depending on the second order moments of
$b_{\h}=\ness{\h}{f_{\mathrm{O}}}$ vanish.

The results show that vast improvements in BER of the I-DJC-DD and
I-DSC-DD receivers with respect to the initialization are obtained,
even for very low $E_b/N_0$ values. As in the case of the DJC-DD and
DSC-DD receivers, joint estimation of the channel weights performs
marginally better than sequential estimation. Both message-passing
receivers clearly outperform the heuristic LMMSE receiver, with
$E_b/N_0$ gains close to 1dB at a BER of 1\%. We explain these gains
by the fact that, contrary to the separate design of the different
modules in the LMMSE receiver, our VMP-SP receivers are analytically
derived based on a global objective function, namely the
region-based free energy. This global design ensures that the
information shared by the
different receiver components is treated correctly, and 
resolves the choice of the appropriate type of information to be
passed from the channel decoder to the other component parts of the
receiver. It is also remarkable that the EM-constrained version of
the I-DJC-DD receiver achieves roughly the same performance as the
non-constrained version. This result seems to indicate that there is
no significant gain to be achieved by computing soft channel
estimates as compared to just point estimates, at least for the
system considered.

\begin{figure}
\centering
\includegraphics[width=0.6\textwidth]{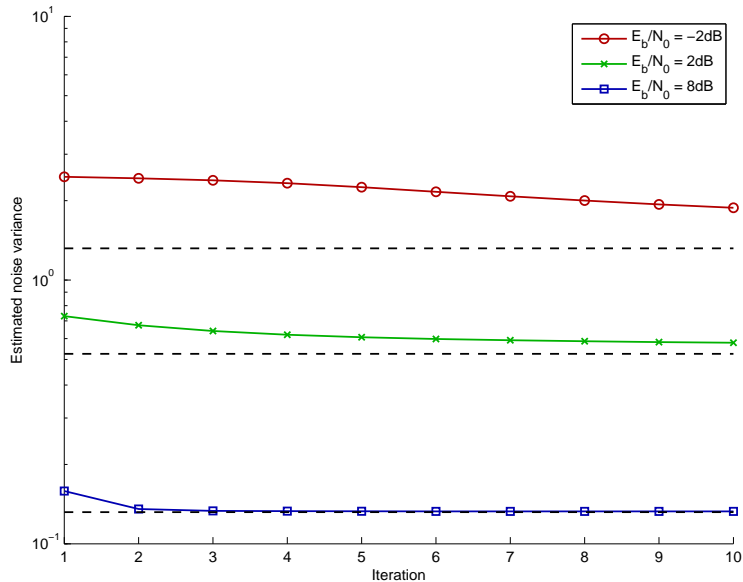}
\caption{Average noise variance estimated by the I-DJC-DD receiver
as a function of the iteration index. 13 pilot symbols are inserted
per OFDM frame. The dashed black lines represent the true noise
variance for each $E_b/N_0$ value.} \label{fig:NoiseEst}
\end{figure}

Another key feature of the I-DJC-DD and I-DSC-DD receivers is the
estimation of the noise precision. This functionality does not only
account for the AWGN, but also includes inaccuracies in the
estimates of the channel weights and data symbols.
Fig.~\ref{fig:NoiseEst} depicts the averaged noise variance estimate
(inverse of the noise precision estimate $\lambda$) provided by the
I-DJC-DD receiver as a function of the iteration index for three
different $E_b/N_0$ levels. The true AWGN variances are also
depicted as dashed black lines. It is apparent from the results that
the behavior of the noise variance estimates (with respect to the
true value) depends heavily on the regime in which the receiver is
operating. For the very low $E_b/N_0$ regime, the receiver
significantly overestimates the noise variance; this is due to the
low accuracy of the channel weights estimates and the large amount
of errors in the estimates of the data symbols obtained. At the
other extreme, for high $E_b/N_0$ values, the estimates of the
channel weights and data symbols become so accurate (as it can be
observed from the low BER values) that the noise variance estimate
rapidly converges to the true AWGN variance, as the contribution of
the estimates' inaccuracies becomes negligible. In the medium
$E_b/N_0$ range, the noise variance estimate slowly converges to a
value larger than the true variance, the difference between both
values depending again on the accuracy of the other parameters'
estimates.

Conceptually, the estimate of the noise precision represents the
amount of `trust' that the algorithm has on the beliefs of the
channel weights and data symbols. With high noise precision values,
the receiver has high confidence on these beliefs, leading to a
rapid convergence towards a stable solution. On the contrary, low
noise precision values will yield slower changes on the beliefs from
one iteration to the next, resulting in a slower convergence rate.


\section{Conclusion}
\label{Sec:Conc}
In this article we have used a hybrid VMP-SP message passing
framework~\cite{Riegler2010,Riegler2011} for the design of iterative
receivers for wireless communication. The framework has been applied
to the factor graph of a generic MIMO-OFDM system. The messages
obtained from the generic derivation have been used to obtain a set
of receiver architectures ranging from computationally simple
solutions to full-scale iterative architectures performing channel
weight estimation, noise precision estimation, MIMO equalization and
channel decoding. The performance of the proposed receivers has been
assessed and compared to state-of-the-art solutions via Monte Carlo
simulations.

A fundamental contribution of this work is the application of a
unified framework that jointly optimizes the receiver architecture
based on a global cost function, namely the region-based variational
free energy. The message-passing scheme used in this work can be
obtained from the equations of the stationary points of a particular
region-based free energy approximation. The resulting algorithm
applies the VMP and SP algorithms to different parts of the graph
and unequivocally defines how the messages of the two respective
frameworks are to be combined. As a result, the hybrid technique
allows for a convenient design of wireless receivers in which the SP
algorithm is used for demodulation and channel decoding and the VMP
algorithm is applied for channel weight estimation, noise covariance
estimation and MIMO equalization. The connection between the
specific receiver component parts is defined by the
message-computation rules, in contrast to other approaches in which
the selection of information to be exchanged among the specific
receiver components is done based on numerical results and/or
intuitive argumentation.

We illustrate the application of the framework by applying it to the
design of receivers in a MIMO-OFDM communications system. From the
factor-graph representing the underlying probabilistic model, a set
of generic messages exchanged between different parts of the model,
represented by sub-graphs, is derived. We choose to split the factor
graph in three main subgraphs corresponding to the channel weights
prior model, the noise precision model and the modulation and coding
constraints. The advantage of this modular approach is that it
enables a scalable, flexible design of the receiver in which the
modification of a specific sub-graph does not modify the processing
performed in other parts of the graph. Thus, a collection of
different receiver architectures can be obtained by applying
different initialization and scheduling strategies. 

In order to assess the performance of the receivers derived with the
proposed framework, we define five specific instances of the generic
message-passing receiver. The particular architectures selected span
from to full-scale iterative schemes, in which the output of the
channel decoder is used to refine the estimates of the channel
parameters and the transmitted symbols, to low-complexity solutions,
in which only pilot symbols are used for channel weight and noise
variance estimation. This particular selection of receiver
architectures serves as an illustration of how the tradeoff between
computational complexity and receiver performance can be adjusted,
with the generic message-passing receiver as a starting point. The
numerical results, obtained via Monte Carlo simulations in a
realistic MIMO-OFDM setup, confirm the effectiveness of the
receivers derived following the hybrid VMP-SP framework. In
particular, the convergence behavior of the receivers tested is
especially remarkable. All receiver instances yield an improved or
equal performance with increasing number of iterations, both in
terms of BER and MSE of the channel weight estimates. We explain
these favorable convergence properties by the use of the unique,
global cost function from which the algorithm is derived. The
estimation of the noise precision parameter, accounting for the
uncertainty on the estimates of the channel weights and transmitted
symbols in addition to the AWGN variance, is another key feature of
the proposed architecture.


\appendices

%

\ifCLASSOPTIONcaptionsoff
  \newpage
\fi

\bibliographystyle{IEEEtran}
\bibliography{IEEEabrv,bibliography}




\end{document}